%% file: main.tex
\documentclass[11pt,letterpaper]{article}
\usepackage[margin=1in]{geometry}
\usepackage{placeins}
\renewcommand{\thetable}{\Roman{table}}
\usepackage[T1]{fontenc}
\usepackage{microtype,graphicx,booktabs,needspace,indentfirst}

\usepackage{xcolor}

\usepackage{amsmath,amssymb,amsthm,mathtools,bm,mathrsfs}
\usepackage{tabularx,array,enumitem,xurl}
\usepackage{tikz,pgfplots}
\pgfplotsset{compat=1.18}
\usepackage[numbers,sort&compress]{natbib}

\usepackage[hidelinks,hyperfootnotes=false]{hyperref}
\usepackage[capitalize,noabbrev]{cleveref}
\graphicspath{{image/}}

\allowdisplaybreaks[2]
\newcounter{algorithm}
\newtheorem{theorem}{Theorem}
\newtheorem{proposition}{Proposition}[section]

\newtheorem{corollary}{Corollary}[section]
\theoremstyle{definition}

\newtheorem{example}{Example}[section]
\theoremstyle{remark}

\theoremstyle{plain}
\crefname{assumption}{Assumption}{Assumptions}
\Crefname{assumption}{Assumption}{Assumptions}
\newcommand{\R}{\mathbb R}
\newcommand{\E}{\mathbb E}

\newcommand{\diag}{\operatorname{diag}}

\title{When Gradient Disagreement Improves Prediction in Looped Networks}
\author{\href{https://orcid.org/0009-0001-6323-5566}{Xinyu Tian}
\quad and \quad \href{https://orcid.org/0000-0003-1300-1451}{Xiaotong Shen}\thanks{Corresponding author: Xiaotong Shen (\href{mailto:xshen@umn.edu}{xshen@umn.edu}).}\\[3pt]
\small School of Statistics, University of Minnesota, Twin Cities}
\date{}

\hypersetup{pdftitle={When Gradient Disagreement Improves Prediction in Looped Networks},pdfauthor={Xinyu Tian and Xiaotong Shen}}
\begin{document}
\makeatletter
\if@filesw\immediate\write\@auxout{\string\citation{v39:reference-style}}\fi
\makeatother
\maketitle
\begin{abstract}
In looped neural networks, the gradient of shared parameters sums
contributions from repeated block applications, obscuring information
about their disagreement. We investigate when this information can improve
prediction and introduce LoopAda, an adaptive optimizer that retains the
summed gradient, pools squared gradient contributions to estimate second moments
for the shared parameters, and selects step sizes using the minibatch loss. An exact moment
decomposition characterizes how gradient dispersion changes the update
direction. For smooth looped predictors fitted by least squares, we
establish sufficient conditions under which LoopAda attains lower expected
excess prediction risk after the initial full-sample update than Adam at
its best fixed scalar rate. The analysis accommodates approximation error.
In an explicit regression model, we establish a uniform separation in
prediction risk over a finite range of iterates and a bounded range of
constant Adam learning rates. This separation persists under small smooth
perturbations of the shared block. Experiments on simulated and real data show that LoopAda can achieve
lower prediction error than Adam.
\end{abstract}


\section{Introduction}\label{sec:introduction-v36}
Computational depth can strengthen neural reasoning without a
corresponding increase in the number of learned
\mbox{parameters}~\cite{saunshi2025,geiping2025recurrent,zhu2025looplm}.
This separation between model size and computation gives parameter sharing
a role in both efficient representation learning and iterative reasoning.
Recursive convolutional networks, Universal Transformers, and
ALBERT~\cite{kim2016recursive,universal,albert} exemplify sharing across
depth. Its computational potential is developed in recurrent networks
that learn iterative procedures~\cite{bansal2022algorithm} and looped
Transformers that execute programmed
algorithms~\cite{giannou2023programmable}. Recent recursive language models
further combine sharing with depth-specific
adaptation~\cite{bae2025recursive} and adaptive
recursion~\cite{bae2025mixture}. The statistical value of this reuse
depends on how the shared transformation is learned.

Parameter sharing creates a distinctive optimization problem: one set of
weights must serve different stages of a computation, and a single update
changes all of those stages simultaneously. Each application of a shared
block acts on a successive intermediate representation and contributes to
the parameter gradient from a different position in the computational
graph. These contributions can differ in magnitude and direction,
reflecting the multiple roles of the same weights. Full backpropagation
sums them to obtain the exact gradient of the training
objective~\cite{werbos1990bptt,pascanu2013difficulty}.
Adam~\cite{kingma2015adam} averages this gradient and its coordinatewise
square to form the update direction, then applies a chosen learning rate.
Its scaling statistic consequently does not distinguish uniformly small
contributions from substantial contributions that cancel. Their dispersion
provides information about the internal structure of the gradient that
the aggregate alone does not retain. We study when this information can
guide adaptive updates toward lower out-of-sample prediction risk.

\begin{figure}[htbp]
\centering
\includegraphics[width=\textwidth]{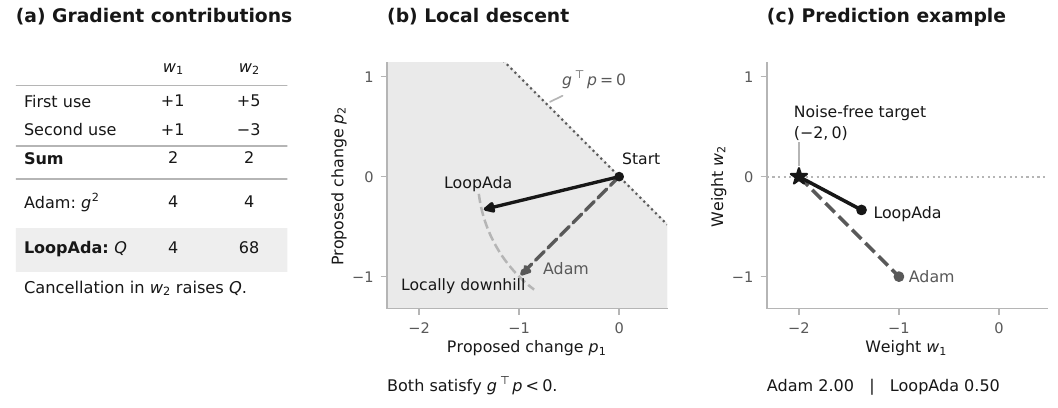}
\caption{Geometry of gradient pooling in a two-weight loop.
(a) Gradient contributions and pooled squares.
(b) Norm-matched descent proposals.
(c) Prediction error relative to the noise-free target.
Proposals use zero initial weights and moments, $\epsilon=0$, and a unit
multiplier before step-size selection.}
\label{fig:loopada-2d}
\end{figure}

We introduce LoopAda, which retains the summed gradient for its first
moment and pools squared contributions for the shared block's second
moment. This construction incorporates disagreement into the adaptive
denominator and can reduce relative movement in coordinates with greater
dispersion. LoopAda uses the minibatch loss to propose a step size and
checks that the proposed step lowers that loss sufficiently.
Figure~\ref{fig:loopada-2d} illustrates how this reallocation can redirect
a descent step and reduce prediction error.

The statistical value of this reweighting depends on where disagreement
occurs relative to signal and nuisance variation. We analyze this
relationship for the first update of smooth \mbox{looped} predictors
approximating nonlinear regression functions. The experiments compare
LoopAda with Adam under squared loss in linear simulations, nonlinear
simulations, and real-data regression cases.

\begin{itemize}
\item \textbf{Adaptive updates for a shared block.} On a common gradient
history, pooled and standard second moments differ by a weighted sum of
squared deviations across block applications. This identity characterizes
the change in update direction (Section~\ref{sec:algorithm-main},
Proposition~\ref{v36:prop:occurrence-geometry}).
\item \textbf{Prediction risk under squared loss.} Conditions on risk
slope and curvature give LoopAda lower expected excess prediction risk
after the initial full-sample update than Adam at its best fixed scalar rate in smooth looped
models (Theorem~\ref{thm:general-risk}). The comparison permits
approximation error and includes a bounded $\tanh$ example. An orthogonal
design admits an exact first-update comparison, with strict advantage
persisting under small changes to the shared block
(Corollary~\ref{cor:general-family}). A two-predictor construction gives
uniform risk separation over a finite range of iterates and a bounded
interval of constant Adam rates (Theorem~\ref{thm:occ-window}).
\item \textbf{Regression studies.} In orthogonal and correlated
designs, the first-update risk ordering changes as disagreement shifts
from nuisance predictors to signal predictors. In nonlinear regression
with repeated measurements, validation-selected LoopAda fits give lower
prediction error than Adam when disagreement is stronger on nuisance
predictors. We also compare test MSE in four real-data regression cases
(Section~\ref{sec:experiments-v36}).
\end{itemize}

\subsection*{Related Work}
Looped Transformers can learn iterative
algorithms~\cite{yang2024learning}, including multistep in-context
gradient descent~\cite{gatmiry2024multistep,chen2025bypassing} and power
iteration with layer normalization~\cite{wu2026power}.
DeepLoop relates alignment across repeated block applications to residual
and initialization scaling~\cite{deeploop}, while
Loop-$\mu$P studies residual scaling and hyperparameter
transfer~\cite{wang2026residual}. Other approaches modify inter-loop
signals for stability~\cite{fu2026fully} or examine the compute--performance
tradeoff of truncated backpropagation~\cite{xiong2026truncate}.

Adaptive preconditioning uses gradient history to determine the geometry
of parameter updates~\cite{duchi2011adagrad,kingma2015adam}.
AMSGrad retains coordinatewise maxima of adaptive second-moment estimates
to address convergence~\cite{reddi2018amsgrad}. AdaBelief constructs its
denominator from deviations of the current gradient from its exponential
moving average~\cite{zhuang2020adabelief}. LoopAda uses the dispersion
across repeated applications of a shared block to construct its adaptive
denominator. This information is available within each gradient evaluation
and is accumulated through the second-moment recursion.

For batch-size invariance, Wang and
Aitchison~\cite{wang2024batchinvariant} average squared microbatch
gradients. PCGrad projects conflicting task gradients before combining
them~\cite{yu2020pcgrad}. LoopAda's contributions arise from repeated
applications within one objective: their sum supplies the first moment
and their squares determine the pooled second moment.

The geometry of an optimization method can affect the selected predictor
and its generalization~\cite{wilson2017marginal}; preconditioning can
change both bias and variance~\cite{amari2021preconditioning}.
For homogeneous neural networks with separable data, implicit-bias
analyses characterize limiting parameter directions of adaptive methods
under exponential and logistic losses~\cite{wang2021implicit}.
Finite optimization trajectories also
provide statistical regularization: early-stopping analyses for kernel
least squares relate the number of gradient steps to prediction
error~\cite{yao2007early,raskutti2014early}.
Our analysis connects the dispersion induced by parameter sharing to
prediction risk under squared loss, including risk separation over a
finite range of iterates.
\section{Looped Networks and Gradient Contributions}\label{sec:loop-setup}
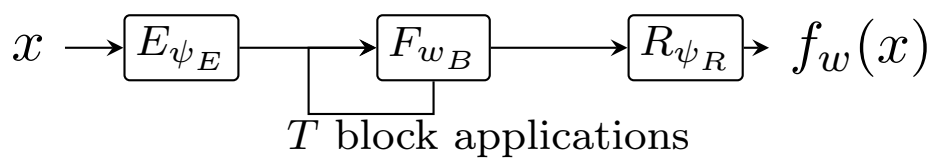
\begin{figure}[htbp]
\centering
\begin{tikzpicture}[x=1.45cm,y=.9cm,>=stealth,box/.style={draw,rounded corners=1pt,minimum width=.66cm,minimum height=.43cm,inner sep=2pt,font=\scriptsize}]
\node (x) at (0,.6) {$x$};
\node[box] (e) at (.7,.6) {$E_{\psi_E}$};
\node[box] (f) at (1.85,.6) {$F_{w_B}$};
\node[box] (r) at (3,.6) {$R_{\psi_R}$};
\node (y) at (3.8,.6) {$f_w(x)$};
\draw[->] (x)--(e);
\draw[->] (e)--(f);
\draw[->] (f)--(r);
\draw[->] (r)--(y);
\draw[->] (f.south) -- +(0,-.24) -- +(-.57,-.24) |- (f.west);
\node[font=\scriptsize] at (2.1,-.07) {$T$ block applications};
\end{tikzpicture}
\caption{Shared-block architecture with $T$ applications of parameters $w_B$.}
\label{fig:shared-loop}
\end{figure}

Let $\mathcal D_n=\{(x_i,y_i)\}_{i=1}^n$ denote the training sample, and let $T$ denote the number of block applications per prediction. The forward recursion is
\[
 \begin{aligned}
 h_0&=E_{\psi_E}(x),\qquad h_t=F_{w_B}(h_{t-1}),\\
 f_w(x)&=R_{\psi_R}(h_T),\qquad t=1,\ldots,T.
 \end{aligned}
\]
The state may include an input sequence and occurrence index. The endpoint group $w_E=(\psi_E,\psi_R)$ contains the encoder and readout, and $w=(w_E,w_B)$ contains all trainable parameters. The same $w_B$ is used at each of the $T$ applications (Figure~\ref{fig:shared-loop}).
{} For a minibatch with index sequence $\mathcal B=(i_1,\ldots,i_m)$, let $\ell$ denote the loss for one observation. The minibatch loss is
\[
 \ell_{\mathcal B}(w)=\frac1m\sum_{j=1}^m\ell\{f_w(x_{i_j}),y_{i_j}\}.
\]
For fixed parameters and a minibatch, let $g_t$ denote the contribution to
a shared tensor's gradient from block application (occurrence) $t$.
The $g_t$ share parameter coordinates, and the chain rule gives the
aggregate gradient $g=\sum_tg_t$~\cite{werbos1990bptt,pascanu2013difficulty}.
We concatenate tensor gradients for parameter groups.

Each LoopAda attempt uses one gradient evaluation and may test several step
sizes. If all trials are rejected, parameters stay fixed and moments are
updated. Theory uses the full training sample; the nonlinear regression simulations
and real-data studies use minibatches of size 64 sampled with replacement. \space
\section{LoopAda}\label{sec:algorithm-main}\label{v36:sec:occurrence-geometry}
Adam forms exponential moving averages of the minibatch gradient and its
coordinatewise square. It divides the bias-corrected first moment by
the square root of the bias-corrected second moment plus a small
constant, then multiplies by a chosen learning rate~\cite{kingma2015adam}.
LoopAda changes two parts of this update. For the shared block, it uses
the squared contributions of the individual block applications to
rescale the update direction. It then uses the minibatch loss to choose
and check a step size (Algorithm~\ref{alg:occurrence-sgd}).
\subsection{Pooled Squared Gradients}
For a shared parameter tensor, the pooled squared-gradient statistic $Q$
has the coordinatewise decomposition
\begin{equation}
\label{eq:occurrence-energy}
 Q:=T\sum_tg_t^{\odot2}
 =g^{\odot2}+T\sum_t(g_t-g/T)^{\odot2}.
\end{equation}
Here $\odot2$ denotes coordinatewise squaring. The second term measures
dispersion of the contributions about their mean across block applications.
The factor $T$ ensures that $Q$ equals the squared aggregate gradient
$g^{\odot2}$ when the contributions are identical.

In Figure~\ref{fig:loopada-2d}(a), contributions $(1,1)$ and $(5,-3)$ have
the same sum but give $Q_1=4$ and $Q_2=68$, respectively. The second
coordinate has the larger denominator at the first update. Norm matching
rescales each tensor's direction to the Frobenius norm of its internal Adam
reference.

At the first update from zero moments, the proposed direction has the
form $p=-Dg$ for a positive diagonal matrix $D$. Its directional derivative
is $g^\top p=-\sum_j D_{jj}g_j^2<0$ whenever $g\ne0$. Positive rescaling by
norm matching preserves the sign. A sufficiently small step in this
direction therefore reduces the minibatch loss
(Figure~\ref{fig:loopada-2d}(b)). At later updates, the running first
moment can change the sign of the directional derivative. LoopAda checks
the derivative and evaluates trial steps on the current minibatch.

Prediction risk depends on movement in signal and nuisance coordinates.
In Figure~\ref{fig:loopada-2d}(c), the training loss is half the squared
distance to $(-2,-2)$, while the noise-free target is $(-2,0)$.
Noise affects only the second coordinate; reducing movement there lowers
squared error relative to the true coefficient vector. Section~\ref{sec:prediction-risk}
gives conditions for the complete update to lower expected excess
prediction risk relative to Adam.

On a common gradient history, let $\widehat m_k$ denote the bias-corrected
first moment. At attempt $k$, write $g_k=\sum_tg_{k,t}$ and
$Q_k=T\sum_tg_{k,t}^{\odot2}$. Define gradient disagreement by
\[
 \mathcal E_k=Q_k-g_k^{\odot2}
 =T\sum_t(g_{k,t}-g_k/T)^{\odot2}.
\]
\begin{proposition}[Gradient dispersion and update direction]
\label{v36:prop:occurrence-geometry}
For one shared tensor, initialize $v_0=q_0=0$. Average $g_k^{\odot2}$ and $Q_k$ with the same decay $\beta_2=.999$, and bias-correct both averages by $1-\beta_2^k$. Then
\begin{equation}\label{v36:eq:moment-excess}
 \widehat q_k-\widehat v_k
 =\frac{1-\beta_2}{1-\beta_2^k}
   \sum_{i=1}^k\beta_2^{k-i}\mathcal E_i\succeq0.
\end{equation}
Here the inequality is coordinatewise. For $\epsilon>0$, define
\begin{equation}\label{v36:eq:coordinate-reallocation}
 \begin{gathered}
 h_{k,j}=\frac{\sqrt{\widehat v_{k,j}}+\epsilon}
                   {\sqrt{\widehat q_{k,j}}+\epsilon}\in(0,1],
 \qquad H_k=\diag(h_{k,j}),\\
 a_k=\frac{\widehat m_k}{\sqrt{\widehat v_k}+\epsilon},
 \qquad u_k=H_ka_k.
 \end{gathered}
\end{equation}
Before any slope fallback, the norm-matched branch uses
\[
 d_k=\frac{\|a_k\|_F}{\|H_ka_k\|_F}H_ka_k,
 \qquad \|d_k\|_F=\|a_k\|_F.
\]
In this branch $d_k=a_k$ exactly when $h_{k,j}$ is constant on the nonzero coordinates of $a_k$; otherwise the direction changes.
\end{proposition}
\begin{proof}
Subtract the two second-moment recursions and unroll them from zero. Equation~\eqref{eq:occurrence-energy} gives \eqref{v36:eq:moment-excess}. Multiplying $a_k$ by $H_k$ gives $u_k$ coordinatewise. Positive diagonal scaling followed by norm matching yields the last assertion. Supplementary Section~\ref{v36:sec:geometry-proof} gives the details.
\end{proof}
For a scalar tensor, norm matching leaves the reference direction
unchanged whenever it is applied. With several coordinates, unequal
scaling can change the direction while preserving its norm. When $T=1$,
the two second-moment estimates coincide, and the fallback and line
search still apply.

Gradient dispersion can be large for coordinates with nonzero true
coefficients. Supplementary Proposition~\ref{v36:prop:signal-damping}
shows that reducing the update in such a coordinate can increase
prediction risk. Supplementary Example~\ref{v36:ex:zero-aggregate}
shows that the pooled squared-gradient statistic can be positive at a
stationary point with a zero aggregate gradient.

Group gradients and directions concatenate the tensor components of $w_E$ or $w_B$. Moment updates and norm matching operate tensorwise; movement caps use group norms. Algorithm~\ref{alg:occurrence-sgd} writes $\ell$ for the current minibatch loss $\ell_{\mathcal B}$.
\begin{figure}[htbp]
\centering
\fbox{\begin{minipage}{.96\textwidth}
\small\raggedright
\refstepcounter{algorithm}\textbf{Algorithm \thealgorithm: LoopAda (diagonal, loss-based)}\label{alg:occurrence-sgd}

\textbf{Input:} shared model parameters $w=(w_E,w_B)$; loop depth $T$; scale $\gamma>0$; exponent $\beta\ge0$. Set $r=T^{-\beta}$ and $\epsilon=10^{-8}$. Initialize $m=v=0$ for every tensor, $q=0$ for shared-block tensors, and $k=0$.

\begin{enumerate}[leftmargin=*,itemsep=1pt,topsep=2pt]
\item Sample a minibatch and evaluate its loss $\ell(w)$. Backpropagate through all $T$ uses, retaining each block contribution $g_t$ and the aggregated gradient $g_B=\sum_t g_t$; obtain endpoint gradient $g_E$.
\item Set $k\leftarrow k+1$. For each parameter tensor, update $m\leftarrow .9m+.1g$, $v\leftarrow .999v+.001g^{\odot2}$ and set $\widehat m=m/(1-.9^k)$, $\widehat v=v/(1-.999^k)$. Let $a=\widehat m/(\sqrt{\widehat v}+\epsilon)$.
\item For each shared-block tensor, update $q\leftarrow .999q+.001T\sum_t g_t^{\odot2}$ and set $u=\widehat m/(\sqrt{q/(1-.999^k)}+\epsilon)$. If $\|u\|_F^2>10^{-60}$, set $d=u\|a\|_F/\|u\|_F$; otherwise set $d=u$. For endpoint tensors, set $d=a$.
\item Compute $s=\langle g_E,d_E\rangle+r\langle g_B,d_B\rangle$. If $s\le0$, set $d=g/(\sqrt{\widehat v}+\epsilon)$ for every tensor and recompute $s$.
\item For $j\in\{E,B\}$, set $c_j=.1\max\{1,\|w_j\|_2\}/\max\{10^{-20},\|d_j\|_2\}$, using group norms. Set $\alpha=\min\{\gamma\ell(w)/\max\{s,10^{-30}\},c_E,c_B/r\}$.
\item For at most eight trials, form $w^+=(w_E-\alpha d_E,w_B-\alpha r d_B)$ and evaluate its loss on the same minibatch. If $s>0$, the trial loss is finite, and $\ell(w)-\ell(w^+)\ge .05\alpha s$, accept $w\leftarrow w^+$ and end the trials. Otherwise halve $\alpha$ and retry.
\item If no trial is accepted, retain $w$. In either case retain the updated moments and repeat from Step 1 until the stopping rule is met.
\end{enumerate}
\end{minipage}}
\end{figure}
\subsection{Preconditioning and Step-Size Selection}
LoopAda keeps Adam's first and second moments of the aggregated gradient, with coefficients $.9$ and $.999$, zero initialization, and $\epsilon=10^{-8}$. Let $\widehat m_k,\widehat v_k$ be their bias-corrected values and $a_k=\widehat m_k/(\sqrt{\widehat v_k}+\epsilon)$ the Adam reference direction~\cite{kingma2015adam}. Adam takes a step of size $\eta$ along $-a_k$, where $\eta$ is the chosen learning rate.
LoopAda uses $a_k$ for the endpoint tensors and as a norm reference for the shared block.
For each shared-block tensor, also maintain
\[
 q_k=.999q_{k-1}+.001T\sum_tg_{k,t}^{\odot2},\qquad
 \widehat q_k=q_k/(1-.999^k).
\]
Put $u_k=\widehat m_k/(\sqrt{\widehat q_k}+\epsilon)$ and use
\[
 d_k=u_k\frac{\|a_k\|_F}{\|u_k\|_F}
 \quad\text{if }\|u_k\|_F^2>10^{-60};
 \qquad d_k=u_k\text{ otherwise}.
\]
The reference $a_k$ is computed from the gradients along the LoopAda
trajectory. When norm matching applies, $\|d_k\|_F=\|a_k\|_F$ for each
tensor. The rule leaves $u_k$ unscaled below the stated threshold and can
replace it through the fallback described below.

The factor $r=T^{-\beta}$ scales the shared-block step relative to the
endpoint step. A nonpositive predicted decrease triggers a fallback
based on the current gradient. The initial step size normalizes the
minibatch loss by the predicted decrease per unit step, with scale
$\gamma>0$ and parameter displacement caps. This construction is related
to stochastic Polyak step sizes, which normalize a loss gap by the
squared gradient norm~\cite{loizou2021sps}.
Backtracking halves the step size after a rejected trial, with at
most eight trials. An accepted step must reduce the minibatch loss by
at least $.05$ times its predicted decrease. This is an Armijo
sufficient-decrease condition~\cite{armijo1966}, evaluated on the current
minibatch as in stochastic line search~\cite{vaswani2019linesearch}.
If all trials are rejected,
the parameters stay fixed and the updated moment estimates are retained.
Theorems~\ref{thm:general-risk} and~\ref{thm:occ-window} analyze this
procedure after one update and over several consecutive updates,
respectively. In the explicit models the endpoints are fixed, and the norms
and inner products for the empty endpoint group are zero.
\section{Prediction Risk in Regression}\label{sec:prediction-risk}
We connect the update geometry to prediction through risk slopes and
curvatures. An orthogonal design makes the role of signal and nuisance
dispersion explicit, and a two-predictor construction extends the
comparison to consecutive updates.

\subsection{A One-Update Guarantee for Smooth Predictors}
Let $Y=f_0(X)+\varepsilon$ with $\E(\varepsilon\mid X)=0$ and
$\E\varepsilon^2<\infty$. The regression function $f_0$ can be any square-integrable function,
including a H\"older-smooth function. We fit a looped predictor $f_w$
by least squares and measure its error against $f_0$ by
\[
 R(w)=\tfrac12\E_X\{f_w(X)-f_0(X)\}^2 .
\]
This risk includes the error of approximating $f_0$ by the fitted
family. The two optimizers use the same family.

Fix common initial weights $w_0$ and zero moments. Both optimizers use
the full training sample for the update. For training
sample $D$, let $p_D$ denote Adam's first direction per unit learning
rate. Let $v_D$ be the first LoopAda direction after norm matching,
relative block scaling, and multiplication by the ratio of training
loss to predicted decrease per unit step. Thus, when the slope
floor and movement caps are inactive and the first trial is accepted,
the first outputs are $w_0-\eta p_D$ and $w_0-\gamma v_D$.
Supplementary Section~\ref{app:directional-risk} gives sufficient
conditions for this first-trial step.

Define the expected risk along the two update paths by
\[
 R_A(t)=\E_D R(w_0-tp_D),\qquad
 R_O(t)=\E_D R(w_0-tv_D).
\]
The slope $a_A=-R_A'(0)$ measures the initial decrease per unit
of Adam rate. Let $a_O\le -R_O'(0)$ be a lower bound on the
corresponding decrease per unit of LoopAda scale.

\begin{theorem}[Lower excess prediction risk after one update]
\label{thm:general-risk}
Suppose the risk curves are twice continuously differentiable on the
stated intervals. Assume $R_A''\ge B_A>0$ on $[0,H]$ for some $H>0$, and
$R_A(\eta)\ge R_A(H)$ for every $\eta>H$. Alternatively, take
$H=\infty$ and require the curvature bound for all $\eta\ge0$.
Suppose $R_O''\le B_O$ on $[0,\Gamma]$, with $B_O>0$, and that
LoopAda accepts its unclipped first trial as above for all
$0<\gamma\le\Gamma$ almost surely. Put
$t_A=\min\{H,\max(a_A,0)/B_A\}$.
Then, for every such $\gamma$,
\begin{equation}\label{eq:directional-gap}
 \inf_{\eta\ge0}R_A(\eta)-R_O(\gamma)
 \ge a_O\gamma-\tfrac12 B_O\gamma^2
     -\bigl(a_A t_A-\tfrac12 B_A t_A^2\bigr).
\end{equation}
If $a_O>0$, the bound is largest at
$\gamma=\min\{\Gamma,a_O/B_O\}$. A positive bound there gives
LoopAda lower risk than Adam at every fixed scalar rate.
\end{theorem}

Supplementary Section~\ref{app:directional-risk} gives the proof.
The slope measures how quickly risk initially falls, and the curvature
limits how far that decrease can continue. The cutoff lets us use
curvature near initialization while retaining the comparison against
all Adam rates. It suffices, for example, that Adam's risk is
nondecreasing beyond the cutoff. Supplementary
Section~\ref{app:bounded-loop-example} verifies the conditions for a
bounded $\tanh$ predictor. Supplementary
Section~\ref{app:holder-loop-example} gives a separate nonlinear
comparison for H\"older targets with positive approximation error.

\subsection{Explicit Conditions on an Orthogonal Design}
Partition $\{1,\ldots,d\}=S\mathbin{\dot\cup}N$, with $d_S=|S|\ge1$ and $d_N=|N|\ge1$. Let $w^\star_j\ne0$ for $j\in S$ and $w^\star_j=0$ for $j\in N$. Write $b_j=|w^\star_j|$. Let $\xi_j$ be independent random signs, with each value in $\{-1,1\}$ equally likely. Define the vector of scaled training responses
\begin{equation}\label{gen:eq:Y}
 Y_j=w^\star_j+\delta_j\xi_j\quad(j\in S),\qquad
 Y_j=\sigma_j\xi_j\quad(j\in N),
\end{equation}
where $\delta_j\ge0$ and $\sigma_j>0$. The training sample consists of inputs $x_j=\sqrt d\,e_j$ and responses $\sqrt d\,Y_j$. A linear predictor $f_w(x)=x^\top w$ therefore has empirical half-squared loss
\begin{equation}\label{gen:eq:loss}
 F_Y(w)=\frac1{2d}\sum_{j=1}^d\{f_w(x_j)-\sqrt d\,Y_j\}^2
       =\frac12\|w-Y\|^2.
\end{equation}
For prediction, the input is uniform on the design points and the
response noise is drawn independently from the same distribution as in
training. The excess prediction risk under half-squared error loss is
\begin{equation}\label{gen:eq:risk}
 \mathcal R(w)=\frac12\|w-w^\star\|^2.
\end{equation}
The prediction risk is the sum of \eqref{gen:eq:risk} and the irreducible
noise term $(\sum_{S}\delta_j^2+\sum_N\sigma_j^2)/2$. When $w$ is estimated
from the training sample, we average $\mathcal R(w)$ over the training
noise to obtain expected excess prediction risk.

The looped network has one trainable tensor, the vector $w$, and fixed
input and output maps. On $x_j$, application $t$ adds the increment
$\sqrt d\,c_{t,j}w_j$ to the linear predictor. The corresponding gradient
contributions satisfy $g_{t,j}=c_{t,j}g_j$ with $\sum_tc_{t,j}=1$ and
$T\sum_tc_{t,j}^2=\kappa_j^2$. The factor $\kappa_j$ quantifies the
dispersion of the contributions for coordinate $j$. A value of one
corresponds to equal contributions. Supplementary equation~\eqref{gen:eq:block}
gives the construction. Use the shared-block scaling $r=T^{-\beta}$
and denominator constant $\epsilon$ from Algorithm~\ref{alg:occurrence-sgd}.
The following sufficient bounds on the coefficients, noise amplitudes,
and optimizer constants give an explicit range of accepted first steps.
\begin{equation}\label{gen:eq:safe}
 \begin{gathered}
 0\le\delta_j\le b_j/10,\quad b_j-\delta_j\ge10^{-4}\quad(j\in S),\\
 \|w^\star\|\le .05,\qquad 1\le\kappa_j\le10\quad(j\in S),\\
 \kappa_j\ge1,\ \sigma_j>0\quad(j\in N),\\
 10^{-12}\le r=T^{-\beta}\le1,\qquad \epsilon=10^{-8}.
 \end{gathered}
\end{equation}
The signal lower bound controls the stabilizer relative to the response;
the amplitude and dispersion bounds ensure that the movement caps remain
inactive over the scale range derived below.

\paragraph{Nonlinear perturbations}
Replace each block increment $u$ by $u+\lambda\varphi(u)$, where
$\lambda\ge0$ and $\varphi$ is continuously differentiable with
$\varphi(0)=\varphi'(0)=0$. The normalized prediction on $x_j$ becomes
$\psi_{\lambda,j}(w_j)$, with
\begin{equation}\label{gen:eq:response-link}
 \psi_{\lambda,j}(s)
 =s+\frac{\lambda}{\sqrt d}\sum_{t=1}^T
              \varphi(\sqrt d\,c_{t,j}s).
\end{equation}
Assume there is a finite $C_\varphi\ge0$ such that
\begin{equation}\label{gen:eq:nonlinear-slope}
 \psi_{\lambda,j}'(s)\ge1-\lambda C_\varphi
 \qquad(j=1,\ldots,d,\ s\in\R,\ \lambda\ge0).
\end{equation}
For small $\lambda$, this slope bound keeps the response sensitive to
changes in the coefficient even far from initialization. It permits a
comparison with Adam over all fixed scalar rates.

The normalized training responses are
$Y_{\lambda,j}=\psi_{\lambda,j}(w_j^\star)+(Y_j-w_j^\star)$, so the noise
in \eqref{gen:eq:Y} is unchanged. Write $\psi_\lambda(w)$ for the vector
of normalized predictions. The excess prediction risk is
$\mathcal R_\lambda(w)=\|\psi_\lambda(w)-\psi_\lambda(w^\star)\|^2/2$.
At $\lambda=0$, the model and risk reduce to their linear forms.

\paragraph{Risk and step size in the linear model}
Define the first-step directions and the loss-based multiplier by
\[
 \begin{gathered}
 p_j(Y)=\frac{Y_j}{|Y_j|+\epsilon},\qquad
 h_j(Y)=\frac{Y_j}{\kappa_j|Y_j|+\epsilon},\\
 a(Y)=\frac{\|Y\|^2}{2Y^\top h(Y)} .
 \end{gathered}
\]
From zero weights and moments, Adam steps to $\eta p(Y)$. On the
unclipped first trial, LoopAda steps to $\gamma a(Y)h(Y)$.
Supplementary \eqref{gen:eq:safe-scale} gives a bound
$\Gamma_{\rm safe}$ such that the first trial is accepted without
clipping or fallback for every $0<\gamma<\Gamma_{\rm safe}$.
Put
\[
 \begin{aligned}
 G_A&=\frac{\{(w^\star)^\top\E p(Y)\}^2}{\E\|p(Y)\|^2},&
 \widetilde m&=\E[a(Y)(w^\star)^\top h(Y)],\\
 \widetilde v&=\E[a(Y)^2\|h(Y)\|^2],&
 M(\gamma)&=\gamma\widetilde m-\tfrac12\gamma^2\widetilde v-\tfrac12 G_A .
 \end{aligned}
\]
The expectations are over the training noise. The scale and Adam's
scalar learning rate are fixed before that noise is drawn.

\begin{corollary}[One-update comparison on the orthogonal design]
\label{cor:general-family}
Assume \eqref{gen:eq:safe}. In the linear model, for
$0<\gamma<\Gamma_{\rm safe}$, the expected risk of best fixed-rate Adam
minus that of LoopAda is exactly $M(\gamma)$.
Let $I\subset(0,\Gamma_{\rm safe})$ be a closed interval on which
$M_I=\min_{\gamma\in I}M(\gamma)>0$.
If \eqref{gen:eq:nonlinear-slope} holds, then for some $\lambda_0>0$ and all
$0\le\lambda\le\lambda_0$, $\gamma\in I$,
\begin{equation}\label{gen:eq:main-bound}
 \E\mathcal R_\lambda\{w_O^+(\gamma,\lambda)\}
 \le \inf_{\eta\ge0}\E\mathcal R_\lambda\{w_A^+(\eta,\lambda)\}-M_I/2 .
\end{equation}
Here $w_O^+$ and $w_A^+$ are the first outputs. LoopAda accepts its
first trial without clipping or fallback for every noise realization.
\end{corollary}

The linear and nonlinear parts are proved in Supplementary
Sections~\ref{app:controller-risk-constants} and~\ref{app:nonlinear-risk}.
The quadratic $M(\gamma)$ gives the range of scales with lower risk than Adam on
this branch, including any effect of the loss-based step size.
Its maximizer $\widetilde m/\widetilde v$ lies in the accepted range
under \eqref{gen:eq:safe}, and the maximum gap is
$(\widetilde m^2/\widetilde v-G_A)/2$.
The slope condition for the nonlinear comparison holds for
$\varphi(u)=\tanh(u)-u$ and, with the block coefficients in Supplementary
\eqref{gen:eq:block}, for positive multiples of odd powers of degree
at least three. Supplementary Section~\ref{app:nonlinear-risk} also gives
a finite cubic example and verifies the first-trial conditions.

\paragraph{A sufficient condition on signal and noise}
An explicit sufficient family combines small signal noise with stronger
gradient disagreement on nuisance coordinates. Under \eqref{gen:eq:safe},
the following conditions give a nonempty interval of accepted scales with
$M(\gamma)>0$ in Corollary~\ref{cor:general-family}.
\begin{equation}\label{gen:eq:family}
 \begin{aligned}
 \text{Counts:}\quad &2d_N\ge d_S,\\
 \text{Signal:}\quad &0\le\delta_j\le .05b_j,\quad 1\le\kappa_j\le1.2
                  &&(j\in S),\\
 \text{Noise only:}\quad &\sigma_j\ge\max_{i\in S}b_i,\quad \kappa_j\ge6
                  &&(j\in N).
 \end{aligned}
\end{equation}
The signal-noise bound controls variation in the loss-based multiplier,
while the dispersion contrast reduces relative movement along nuisance
coordinates. Together, the conditions imply
$\widetilde m^2/\widetilde v>G_A$, giving an interval of lower risk.
The count condition permits one nuisance predictor for every two signal
predictors. Supplementary Section~\ref{app:dispersion-conditions} proves
the sufficiency of~\eqref{gen:eq:family}.

The linear result also covers the model with two predictors in Supplementary
Theorem~\ref{thm:occ-first-risk}. Supplementary Corollary~\ref{gen:cor:original}
recovers its step interval and risk margin.

\begin{figure}[htbp]\centering
\includegraphics[width=\columnwidth]{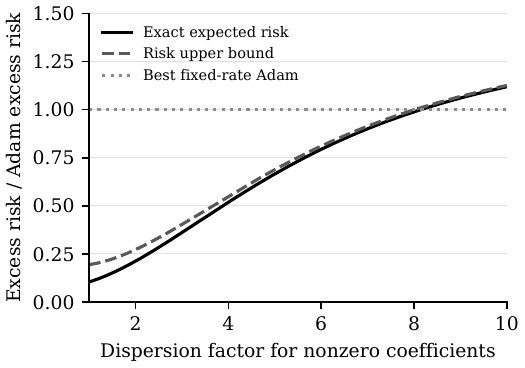}
\caption{Excess-risk ratio as signal-coordinate dispersion increases.
Solid: exact risk at $\gamma_c$ in Supplementary~\eqref{gen:eq:conservative-scale};
dashed: its upper bound; horizontal reference: best fixed-rate Adam.
Other parameters are fixed at the nine-predictor setting in Supplementary
Section~\ref{app:nine-predictor}.}
\label{fig:general-alignment}\end{figure}

Supplementary Section~\ref{app:nine-predictor} evaluates the expected
risks at the boundary $2d_N=d_S$ and derives the scale interval for a
nine-predictor example from the roots of $M(\gamma)$. These examples
illustrate the magnitude and persistence across scales of the strict
advantage. Figure~\ref{fig:general-alignment} shows how the comparison
changes as gradient dispersion increases on the signal coordinates.
\subsection{Prediction Risk over Consecutive Updates}
Over consecutive updates, the moment estimates depend on the evolving
parameter trajectory. We give a concrete comparison in an orthogonal
model with two predictors, using the shared-block perturbation of
Corollary~\ref{cor:general-family}. The parameter box, iteration window,
and Adam rate interval specify the construction; its risk gaps follow
from uniform bounds. This finite-iterate analysis complements analyses of
early stopping as statistical regularization in least-squares
learning~\cite{yao2007early,raskutti2014early}.

In the linear model, the training loss and excess prediction risk are
\begin{equation}\label{eq:window-model-main}
 \begin{aligned}
 F(w)&=\tfrac12\{(w_1-\mu)^2+(w_2-\chi\sigma)^2\},\\
 \mathcal R(w)&=\tfrac12\{(w_1-\mu)^2+w_2^2\}.
 \end{aligned}
\end{equation}
The dispersion factors for the two predictors are $1,\kappa$, and
$\chi=\pm1$ is the training noise sign. Let $K$ count optimizer attempts.
Write $\theta=(\mu,\sigma,\kappa,\gamma)$ and define
\begin{equation}\label{eq:occ-window-box}
 \begin{split}
 \theta_0&=(.01,.1,10,1/256),\\
 \mathcal P&=\{\theta:|\theta_j/\theta_{0,j}-1|\le10^{-5},\ j=1,\ldots,4\}.
 \end{split}
\end{equation}
All four parameters may vary independently within this box. Use the
block in Supplementary \eqref{eq:occ-risk-block} with $T=8$,
$r=8^{-1/2}$, and $\epsilon=10^{-8}$. For a fixed continuously
differentiable $\varphi$ with $\varphi(0)=\varphi'(0)=0$, replace each
block increment $u$ by $u+\lambda\varphi(u)$. The normalized
training response is $(\psi_{\lambda,1}(\mu),\chi\sigma)$, and the
risk is $\mathcal R_\lambda(w)=\|\psi_\lambda(w)-\psi_\lambda(\mu,0)\|^2/2$.
Start both optimizers at zero weights and
moments, keep the endpoints fixed, and use the full training pair
at each update. Write $w_{K,\lambda}^O$ and $w_{K,\lambda}^A(\eta)$
for their outputs. Adam uses the constant scalar block rate $\eta$,
coefficients $.9,.999$, and the same $\epsilon$.

\Needspace{10\baselineskip}
\begin{theorem}[Uniform risk separation over a finite range of iterates]
\label{thm:occ-window}
For each such $\varphi$ there exists $\bar\lambda>0$ such that for
$0\le\lambda\le\bar\lambda$, every parameter vector in $\mathcal P$,
and either noise sign,
\begin{equation}\label{eq:occ-window-bound}
 \max_{6\le j\le16}\mathcal R_\lambda(w_{j,\lambda}^O)
 \le .65\!
 \inf_{\substack{6\le k\le16\\0\le\eta\le.1}}
 \mathcal R_\lambda\{w_{k,\lambda}^A(\eta)\}.
\end{equation}
The absolute risk gap is at least $7.9\times10^{-6}$.
LoopAda accepts all sixteen first trials without clipping or fallback.
For $\varphi(u)=u^3$, one may take $\bar\lambda=10^{-3}$.
\end{theorem}

The uniform bound permits distinct iteration indices for the two methods
and any constant Adam rate in the stated interval. It also holds for
data-dependent choices within these ranges and in expectation over the
noise sign.

\begin{figure}[htbp]\centering
\includegraphics[width=\columnwidth]{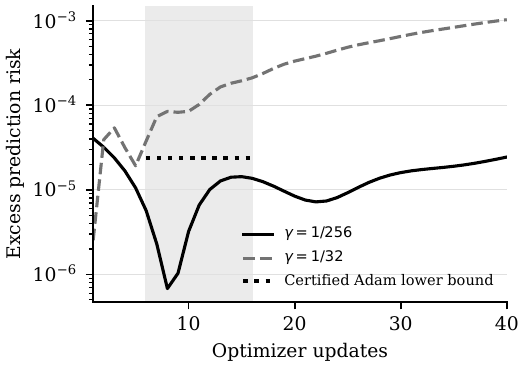}
\caption{Prediction risk over successive updates in the two-predictor model
at the central parameter values. Shading marks the range covered by
Theorem~\ref{thm:occ-window} for scales near $1/256$; the dotted segment is
Adam's uniform lower bound.}
\label{fig:finite-window}\end{figure}

The proof is in Supplementary Section~\ref{app:occ-window}:
Sections~\ref{app:window-invariants}--\ref{app:window-linear} establish the
linear bounds, and Section~\ref{app:nonlinear-window} proves the nonlinear
and cubic cases. In the linear case, the ordering
can reverse at update two at the larger LoopAda scale $1/32$.
If both methods interpolate the same training responses in that case,
their prediction risks coincide. Figure~\ref{fig:finite-window} shows the
theorem's iteration window and the effect of scale in the linear model.

\section{Experiments}\label{sec:experiments-v36}
We compare LoopAda and Adam in three squared-loss studies. The linear
simulation varies which predictors carry signal to examine the first
update. The nonlinear simulation compares validation-selected fits from
repeated measurements. The real-data cases compare test MSE on held-out
groups.

\subsection{Signal Placement in Linear Regression}\label{sec:signal-placement}
A fixed depth-eight loop estimates a linear regression function with eight
predictors. Four coordinates have dispersion factor $\kappa_j=1.1$ and
four have factor $\kappa_j=8$. In each setting, four true regression
coefficients equal $.01$ and the remaining four equal zero. We vary the
number of nonzero coefficients in the high-dispersion group, holding the
network and coefficient magnitudes fixed. In the orthogonal design,
noise amplitudes are assigned by whether a coefficient is zero. In the
correlated design, the response-noise variance is fixed. Each optimizer
starts at zero and takes one full-sample update. LoopAda uses
scale $.45$ throughout. Adam uses the fixed scalar learning rate that
minimizes expected excess prediction risk in the orthogonal design. In
the correlated design we estimate that rate on independent simulated
samples. Supplementary Section~\ref{sec:signal-placement-supp} gives the
designs and rate selection.

\begin{figure}[htbp]
\centering
\includegraphics[width=\columnwidth]{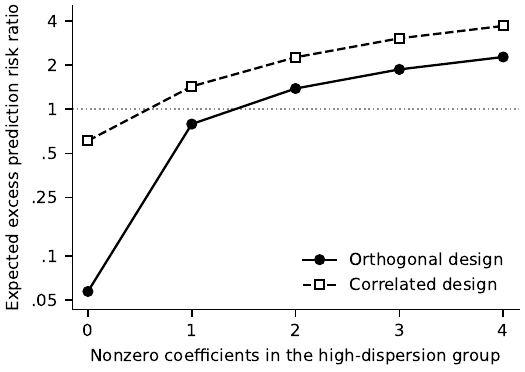}
\caption{First-update excess-risk ratio (LoopAda/Adam) as signal shifts
to the high-dispersion group. The dotted line denotes equal risk.
Risks average over all 256 orthogonal noise patterns or 4,096
correlated-design samples.}
\label{fig:signal-placement}
\end{figure}

\begin{figure}[htbp]
\centering
\includegraphics[width=\textwidth]{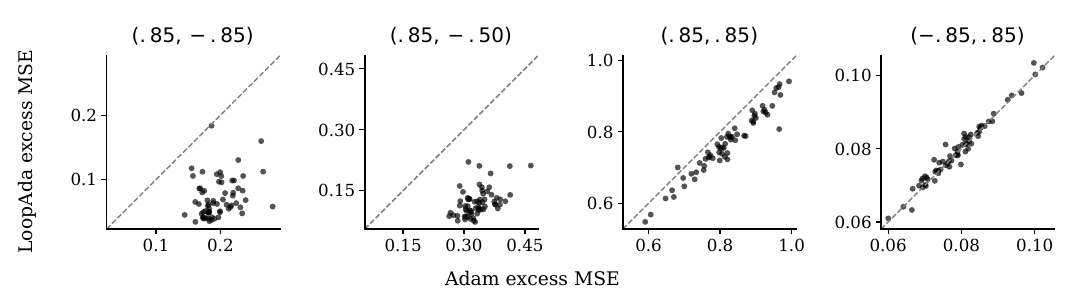}
\caption{Paired excess MSEs for LoopAda and Adam in nonlinear regression.
Each point represents one of 64 paired replications at a maximum of
480 update attempts. Panel titles give the signal and nuisance
autocorrelations $(\rho_S,\rho_N)$. Points below the diagonal favor
LoopAda. Both axes use the same scale within a panel.}
\label{fig:nonlinear-regression}
\end{figure}

The risk ordering reverses as disagreement shifts from nuisance to signal
coordinates. In the orthogonal design, the ratio of expected excess
prediction risks is $.057$ when all high-dispersion coordinates have zero true coefficients
(Figure~\ref{fig:signal-placement}). The ratio exceeds one as more nonzero
coefficients are assigned to that group and reaches $2.275$ when the
group contains all four nonzero coefficients. The reversal persists
when LoopAda uses its best scale on the accepted, unclipped first-trial
branch for each orthogonal placement. The optimized ratios at the two
endpoints are $.037$ and $1.963$. In the correlated Gaussian design,
with LoopAda's scale fixed, the Monte Carlo estimates range from
$.610$ to $3.706$ at the same endpoints. 
\subsection{Nonlinear Regression with Repeated Measurements}
\label{sec:nonlinear-regression}
This study compares validation-selected LoopAda and Adam fits for
nonlinear regression with repeated measurements. Each observation contains eight
measurements on 64 independent predictors. The measurements on four signal
predictors persist over time, while those on 60 nuisance predictors
alternate. Each predictor follows a stationary Gaussian AR(1) process,
with lag-one correlation $.85$ for signal predictors and $-.85$ for
nuisance predictors. The conditional mean is a nonlinear function of the
four signal trajectories; responses add independent Gaussian noise of
unit variance.

Both methods fit the same network. Given the current measurement
vector $x_t$, the shared block updates the state by
\[
 h_t=h_{t-1}+\tfrac18\tanh(w\odot x_t+a\odot h_{t-1}),
 \qquad h_0=0.
\]
The prediction sums the final state coordinates and adds an intercept.
The input weights $w$, feedback weights $a$, and intercept are trainable,
with the same initialization law for every predictor. Signal and nuisance
labels enter the data generator and gradient summaries.

Paired fits share 128 training observations, initialization, and minibatch
sequences. Excess MSE is evaluated against the true conditional mean on
4,096 new inputs per replication. With a maximum of 120 update attempts,
32 development replications, and 128 paired evaluation replications,
the mean excess MSE is $.074$ for LoopAda and $.205$ for Adam
(Supplementary Section~\ref{sec:nonlinear-regression-supp}).

\paragraph{Training horizon and temporal correlation}
We vary the signal and nuisance autocorrelations, denoted by $\rho_S$
and $\rho_N$, and the maximum number of update attempts $H$. A LoopAda
attempt counts toward $H$ even if the step is rejected. Each method is
tuned separately in each setting on 16 development replications and
evaluated on 64 new paired replications. Within each replication, the
checkpoint is chosen using independent noisy validation responses. The
budgets share data, initialization, and minibatch prefixes.

\begin{table}[t]
\centering\small
\caption{Mean excess MSE for LoopAda and Adam in nonlinear regression}
\label{tab:nonlinear-strong-baselines}
\setlength{\tabcolsep}{3pt}
\begin{tabular}{@{}crrr@{}}
\toprule
$(\rho_S,\rho_N)$ & $H$ & LoopAda & Adam\\
\midrule
$(.85,-.85)$ & 120 & $.0712\,(.0038)$ & $.1963\,(.0035)$\\
$(.85,-.85)$ & 480 & $.0712\,(.0038)$ & $.1963\,(.0035)$\\
$(.85,-.85)$ & 1920 & $.0712\,(.0038)$ & $.1963\,(.0035)$\\
$(.85,-.50)$ & 480 & $.1212\,(.0043)$ & $.3337\,(.0048)$\\
$(.85,.85)$ & 480 & $.7708\,(.0109)$ & $.8221\,(.0117)$\\
$(-.85,.85)$ & 480 & $.0795\,(.0011)$ & $.0790\,(.0011)$\\
\bottomrule
\end{tabular}
\par\smallskip
\begin{minipage}{\linewidth}\footnotesize
Entries give mean excess MSEs with Monte Carlo standard errors in
parentheses, based on 64 paired evaluation replications.
Adjusted paired bootstrap intervals appear in Supplementary
Table~\ref{tab:nonlinear-followup-ratios}.
\end{minipage}
\end{table}

With signal autocorrelation $.85$ and nuisance autocorrelation $-.85$,
LoopAda's mean excess MSE at $H=480$ is $63.7\%$ lower than Adam's,
with an adjusted $95\%$ interval of $57.6\%$ to $69.0\%$ for the
reduction (Table~\ref{tab:nonlinear-strong-baselines} and Supplementary
Table~\ref{tab:nonlinear-followup-ratios}). Validation selects the same
early fits for these correlations at all three budgets, yielding
identical mean errors across budgets.
At $H=480$, the reduction is about $6\%$ when both correlations are
positive. When the correlation ordering is reversed, Adam has the
smaller point estimate and the adjusted interval includes one. This
setting also changes the variance of the true conditional mean
(Supplementary Section~\ref{sec:nonlinear-followup-supp}).
Figure~\ref{fig:nonlinear-regression} shows the paired errors at $H=480$
in all four correlation settings.

\subsection{Real-Data Regression Cases}\label{sec:real-regression}\label{sec:occurrence-nn-results}
We examine Portuguese student grades~\cite{uci_student_performance},
CT slice location~\cite{uci_ct_slices}, community violent-crime
rates~\cite{uci_communities_crime}, and diabetes
progression~\cite{diabetes_original_raw_mirror} under squared loss.
Test MSE is measured on the original response scale. For a common data
distribution, population-MSE differences equal twice differences in
$R(w)$; MSE ratios also contain irreducible response variation.

LoopAda and Adam fit the same width-32 $\tanh$ network at depth $T=8$, with trainable
encoder, shared block, and readout. Each fit permits 1,200 update
attempts with minibatches of size 64. Five outer group folds hold out
profile keys, released patient identifiers, states, or individual rows,
respectively. Four paired development repetitions select each method's
rate and block scaling independently; four fresh repetitions per fold
supply out-of-fold predictions. We select the checkpoint on the stopping set
and estimate standardization parameters from the fitting set. Methods share partitions,
initialization, and minibatch streams. We average individual squared
errors rather than ensemble predictions. Supplementary
Section~\ref{sec:real-regression-supp} specifies data preparation,
selection, and the paired uncertainty analysis.

\begin{table}[t]
\centering\small
\caption{Mean test MSE for LoopAda and Adam in four regression cases}
\label{tab:real-regression}
\setlength{\tabcolsep}{4.5pt}
\begin{tabular}{@{}lrr@{}}
\toprule
Case & LoopAda & Adam\\
\midrule
Portuguese grades & $2.74627$ & $2.91977$\\
CT slice location & $46.4449$ & $50.4929$\\
Communities and Crime & $.0242271$ & $.0245137$\\
Diabetes & $3200.81$ & $3198.46$\\
\bottomrule
\end{tabular}
\par\smallskip
\begin{minipage}{\linewidth}\footnotesize
Original-scale squared losses are averaged over four fits per outer
fold. Pointwise 95\% paired bootstrap intervals appear in Supplementary
Table~\ref{tab:real-regression-ratios}.
\end{minipage}
\end{table}

LoopAda achieved lower mean test MSE than Adam in three of the four
regression cases (Table~\ref{tab:real-regression}). The largest reduction
was for CT slice location ($8.02\%$), followed by Portuguese grades
($5.94\%$) and Communities and Crime ($1.17\%$). On diabetes, LoopAda's
mean test MSE was $0.07\%$ higher. The pointwise $95\%$ interval for the
LoopAda-to-Adam MSE ratio lay below one for Portuguese grades
($[0.887,0.990]$) and included one for the other three cases.
\section{Discussion}\label{sec:discussion-main}
Theorem~\ref{thm:general-risk} gives a one-update risk advantage under
directional conditions permitting bounded nonlinear predictors.
Corollary~\ref{cor:general-family} gives an exact gap on an orthogonal
linear design and extends the comparison to small nonlinear perturbations.
In a two-predictor model, Theorem~\ref{thm:occ-window} separates risks uniformly over
a finite range of iterates and a bounded range of constant Adam rates,
including an
explicit range of cubic perturbations.

In the linear simulations, LoopAda's first-update advantage reverses when
gradient disagreement shifts from nuisance predictors to signal predictors. In nonlinear regression with
signal autocorrelation $.85$ and nuisance autocorrelation $-.85$,
disagreement at initialization accounts for about $98\%$ of pooled
gradient energy on nuisance predictors and $8\%$ on signal predictors
(Supplementary Section~\ref{sec:nonlinear-followup-supp}).

LoopAda differs from Adam in both the second-moment calculation and
the step-size rule. To assess pooling separately, the supplementary
control squares the summed gradient, as Adam does, in place of pooling
the squared contributions from individual uses of the block. The
control keeps LoopAda's rule for proposing and checking the step size
using the minibatch loss. With separate tuning, pooling lowers mean
excess MSE by about $65\%$ in the setting with negatively correlated
nuisance measurements described above. With equal correlations the reduction
is about $6\%$. When the correlation ordering is reversed, the
adjusted interval includes one (Supplementary
Section~\ref{sec:nonlinear-followup-supp}).

Interpolating the same responses gives both methods equal risk in the
linear two-predictor model. A finite line search can reject every trial at
a nonstationary point, even for smooth strongly convex losses
(Supplementary Section~\ref{sec:supp-stall}). Analysis over longer trajectories must
therefore connect gradient dispersion, scale selection, and stopping.

Future work includes risk guarantees for data-dependent \mbox{scales}, identifying
noise-sensitive parameters from dis\-agree\-ment, and studying these
relationships in larger recurrent networks, tied-weight Transformers, and
convolutional layers.
\section*{Acknowledgment}
This work was supported in part by the National Science Foundation (NSF) under Grant DMS-2513668, by the National Institutes of Health (NIH) under Grants R01AG069895, R01AG065636, and R01AG074858, and by the Minnesota Supercomputing Institute.

The authors utilized AI-assisted tools for coding support and prose
refinement. Additionally, selected arithmetic claims in the numerical
studies were formally verified using the Lean proof assistant. The authors
assume full responsibility for the integrity of the manuscript, proofs,
code, and final results.

\label{main:end}
\clearpage
\pdfbookmark[0]{Supplementary Material}{supplement-start}
\begin{center}\LARGE\bfseries Supplementary Material\end{center}
\setcounter{figure}{0}
\setcounter{table}{0}
\renewcommand{\thetable}{\arabic{table}}
\renewcommand{\theHfigure}{supp.\arabic{figure}}
\renewcommand{\theHtable}{supp.\arabic{table}}
\numberwithin{equation}{section}
\counterwithin{theorem}{section}
\input{supplement}
\FloatBarrier
\bibliographystyle{IEEEtranN}
\bibliography{bib/references}
\end{document}

%% file: supplement.tex
This supplement presents the proofs, interval bounds, algorithmic examples,
and experimental methods and results supporting the main paper.

\pdfbookmark[1]{Guide to the Proofs}{proof-guide}
\section*{Guide to the Proofs}
The table gives the starting page of each proof for the main-text results.

\begin{center}
\small
\renewcommand{\arraystretch}{1.25}
\begin{tabularx}{\linewidth}{@{}lXr@{}}
\toprule
Main-text result & Proof location & Page\\
\midrule
Proposition~\ref{v36:prop:occurrence-geometry}
 & Gradient dispersion and update direction
   (Section~\ref{v36:sec:geometry-proof})
 & \pageref{v36:sec:geometry-proof}\\
Theorem~\ref{thm:general-risk}
 & Directional risk comparison (Section~\ref{app:directional-risk})
 & \pageref{proof:theorem-one}\\
Corollary~\ref{cor:general-family}, linear case
 & Accepted first update and exact risk gap
   (Section~\ref{app:controller-risk-constants})
 & \pageref{proof:corollary-linear}\\
Corollary~\ref{cor:general-family}, nonlinear case
 & Persistence of the risk gap (Section~\ref{app:nonlinear-risk})
 & \pageref{proof:corollary-nonlinear}\\
Sufficient family in~\eqref{gen:eq:family}
 & Verification of a positive risk margin
   (Section~\ref{app:dispersion-conditions})
 & \pageref{proof:sufficient-family}\\
Theorem~\ref{thm:occ-window}
 & Uniform risk separation over a finite range of iterates
   (Section~\ref{app:occ-window})
 & \pageref{app:occ-window}\\
\bottomrule
\end{tabularx}
\end{center}

The proof of Corollary~\ref{cor:general-family} first identifies the
accepted update and its exact linear risk, then extends the comparison
under smooth perturbations. The proof of Theorem~\ref{thm:occ-window}
proceeds from moment invariants and first-trial acceptance to uniform
linear risk bounds, followed by the nonlinear and cubic cases.

Section~\ref{app:occurrence-risk} develops the two-predictor construction
and supplementary results, which carry their own appendix numbering.
Sections~\ref{app:bounded-loop-example}--\ref{app:holder-loop-example}
give nonlinear examples, and Section~\ref{app:nine-predictor} gives
explicit risk calculations. Section~\ref{sec:experimental-methods}
contains the experimental methods and results.
\clearpage
\appendix
\section{Gradient Dispersion and Shared-Block Updates}
\label{v36:app:occurrence-geometry}

\subsection{Proof of Proposition~\ref{v36:prop:occurrence-geometry}}
\label{v36:sec:geometry-proof}
The argument combines the pooled-gradient identity, the second-moment
recursion, and the norm-matching rule.

\paragraph{Gradient decomposition.}
Use the notation of Section~\ref{v36:sec:occurrence-geometry}. With predictions and responses normalized by the square root of sample size, let $\Phi$ be the prediction vector and $e$ its residual. Under half-squared loss, $J_t$ is the contribution of block use $t$ to the Jacobian of $\Phi$, so $g_t=J_t^\top e$. Write $\bar J=T^{-1}\sum_tJ_t$ and $\bar g=g/T$. The identity $\sum_t(g_t-\bar g)=0$ proves the coordinatewise decomposition \eqref{eq:occurrence-energy}, and
\[
 \sum_t\|g_t-\bar g\|^2
 =e^\top\!\left\{\sum_t(J_t-\bar J)(J_t-\bar J)^\top\right\}e.
\]
Restricting $J_t$ to the columns of one tensor gives its contribution to the dispersion.

\paragraph{Second-moment difference.}
Subtraction of the second-moment recursions gives
\[
 q_k-v_k=\beta_2(q_{k-1}-v_{k-1})+(1-\beta_2)\mathcal E_k,
 \qquad q_0-v_0=0.
\]
Unrolling and dividing by $1-\beta_2^k$ gives \eqref{v36:eq:moment-excess}. Since every $\mathcal E_k$ is nonnegative coordinatewise, $\widehat q_k\succeq\widehat v_k\succeq0$. The positive denominator constant implies $0<h_{k,j}\le1$, including coordinates with zero second moment, and direct cancellation gives $H_ka_k=u_k$.

\paragraph{Norm matching.}
In the active norm-matching branch, $H_ka_k\ne0$ and hence $a_k\ne0$. The equation $d_k=a_k$ is equivalent to
\[
 H_ka_k=\frac{\|H_ka_k\|_F}{\|a_k\|_F}a_k,
\]
so $h_{k,j}$ must be constant on the support of $a_k$; this condition also suffices. Positive diagonal scaling preserves the orientation of any collinear vector, so equality of the two norms proves the direction statement.\hfill\qedsymbol

If the occurrence Jacobians agree initially and the moments start at zero, then $\widehat q_1=\widehat v_1$ and $u_1=a_1$. Both normalization branches return $a_1$ before any slope fallback, including when it is zero. Vanishing disagreement at a later single update need not erase a difference already accumulated in the moments.

\subsection{Prediction Directions and Scalar Rescaling}
For fixed endpoints, a shared block displacement satisfies
\[
 \Phi(\theta+\Delta\theta)-\Phi(\theta)
 =S\Delta\theta+o(\|\Delta\theta\|),\qquad S=\sum_tJ_t.
\]
Reweighting that displacement cannot enlarge its first-order image beyond $\operatorname{range}(S)$. With trainable endpoints, the first-order image is contained in $\operatorname{range}[J_E,S]$, where $J_E$ is the endpoint Jacobian.

For a prescribed gradient history $g'_i=Tg_i$ and zero initial moments, Adam has $\widehat m'_k=T\widehat m_k$ and $\widehat v'_k=T^2\widehat v_k$. Therefore
\begin{equation}\label{v36:eq:adam-history-scaling}
 \frac{\widehat m'_k}{\sqrt{\widehat v'_k}+\epsilon}
 =\frac{\widehat m_k}{\sqrt{\widehat v_k}+\epsilon/T}.
\end{equation}
This identity describes a scalar rescaling of a common prescribed gradient history.

\section{Two-Predictor Construction and Supplementary Results}
\label{app:occurrence-risk}
The construction and risk calculations use real arithmetic and the acceptance rule of Algorithm~\ref{alg:occurrence-sgd}, including $\epsilon=10^{-8}$. The encoder and readout are fixed, so the endpoint parameter group is empty.

\subsection{Linear Regression with Two Predictors}\label{sec:occurrence-risk-main}
The optimizer in Section~\ref{sec:algorithm-main} reallocates movement between coordinates. We evaluate this reallocation by prediction risk on new observations: fitting noisy training responses more closely can increase prediction risk. Consider linear regression with two predictors and fixed input and output maps. A fixed encoder and readout and one repeated linear block implement $f_w(x)=x^\top w$, where $w=(w_1,w_2)$ forms one trainable tensor. With $e_1,e_2$ the coordinate unit vectors, the training inputs are $x_i=\sqrt2e_i$ and the responses are $y=\sqrt2(\mu,\xi)$, where $\mu>0$ is signal strength and the observation noise $\xi$ equals $\sigma$ or $-\sigma$ with equal probability. The training loss and excess prediction risk are
\begin{equation}\label{eq:occ-risk-model-main}
 \begin{aligned}
 F(w)&=\tfrac12\{(w_1-\mu)^2+(w_2-\xi)^2\},\\
 \mathcal R(w)&=\tfrac12\{(w_1-\mu)^2+w_2^2\}.
 \end{aligned}
\end{equation}
Here population inputs are uniform on the design pair; the first response equals $\sqrt2\mu$, and the second has an independent symmetric noise response. Thus population half-squared loss is $\mathcal R(w)+\sigma^2/2$.

Write $g^{(j)}$ and $g_t^{(j)}$ for coordinate $j$ of the aggregate and occurrence gradients. Section~\ref{sec:occurrence-block-construction} constructs an identical linear block at every use for which
\[
 g_t^{(1)}=g^{(1)}/T,\qquad
 T\sum_t(g_t^{(2)})^2=\kappa^2(g^{(2)})^2,\quad \kappa>1.
\]
The signal contributions are equal, while the noise-only coordinate's pooled squared-gradient statistic is $\kappa^2$ times its squared aggregate gradient. An auxiliary hidden state changes across applications of the identical block, producing different coordinate-2 contributions. The construction allows any fixed $\kappa>1$, independently of $T$, specified by the architecture before the training noise is drawn. The architecture thus assigns stronger dispersion to the noise coordinate.

\Needspace{8\baselineskip}
\begin{theorem}[A first-update prediction-risk advantage]\label{thm:occ-first-risk}
Consider this shared linear model with any fixed dispersion factor $\kappa>1$, $T\ge2$, $10^{-4}\le\mu\le.05$, $\sigma>0$, and $10^{-12}\le r=T^{-\beta}\le1$. Initialize $w=0$ and all moments at zero, use the full training pair, and keep $\epsilon=10^{-8}$ and all safeguards of Algorithm~\ref{alg:occurrence-sgd}; the endpoint parameter group is empty. There is a nonempty interval of positive $\gamma$, given in \eqref{eq:occ-risk-gamma}, on which LoopAda accepts its first trial without clipping or fallback and satisfies
\begin{equation}\label{eq:occ-risk-comparison-main}
 \mathbb E\mathcal R(w_O^+)
 <\inf_{\eta\ge0}\mathbb E\mathcal R\{w_A^+(\eta)\}.
\end{equation}
Here $w_A^+(\eta)$ is Adam's first update from the same batch and zero moments, with any scalar learning rate $\eta$. The interval is independent of the realized noise sign. Section~\ref{sec:occurrence-first-proof} gives its endpoints, a positive risk margin, and the full proof.
\end{theorem}

From zero moments, Adam's first update points along a fixed ray; its scalar learning rate determines the distance along that ray, including at the population-risk minimum. LoopAda changes the relative movement of the two coordinates. The pooled denominator gives additional damping in the noise-only coordinate relative to the Adam reference, limiting movement toward the noisy response while retaining progress on the signal. This advantage depends on aligning high dispersion with noise; assigning it to signal can reverse the comparison. The result concerns the first update with fixed endpoints and zero initial moments. Section~\ref{sec:occurrence-nn-results} studies subsequent training with nonlinear blocks, trainable endpoints, and validation-based selection empirically.

For $T=8$, $\mu=.01$, $\sigma=.1$, $\kappa=10$, $r=8^{-1/2}$, and $\gamma=1/32$, the first-update excess risk is $2.5360\times10^{-6}$, compared with $2.5000\times10^{-5}$ for Adam at its best fixed scalar rate. Section~\ref{sec:occurrence-first-proof} verifies the step conditions and gives the exact comparison.

\paragraph{Dispersion on the signal coordinate}
The same architecture also gives a converse comparison. Keep the squared-gradient multipliers $1$ and $\kappa^2$ on the two coordinates, but put noise in the first coordinate and signal in the second. Proposition~\ref{v36:prop:signal-damping} proves that the smallest prediction risk along the occurrence direction is then strictly larger than the smallest risk along Adam's direction. This covers the complete first update: norm matching, block scaling, caps and backtracking only select a nonnegative distance on that direction, and a rejected update returns its origin. The allocation of signal and noise determines whether the occurrence direction lowers risk. Main-text identity~\eqref{eq:occurrence-energy} specifies dispersion while leaving this allocation unrestricted.

\subsection{An Identical Repeated Block with Prescribed Dispersion}
\label{sec:occurrence-block-construction}
For $T\ge2$ and $\kappa>1$, define
\[
 b_\kappa=\sqrt{\frac{12(\kappa^2-1)}{T^2(T^2-1)}},
 \qquad
 a_\kappa=\frac1T-\frac{T-1}{2}b_\kappa.
\]
Use encoder $x\mapsto(x_1,x_2,0,0)$, readout $(x_1,x_2,z,h)\mapsto h$, and repeat
\begin{equation}\label{eq:occ-risk-block}
 F_w(x_1,x_2,z,h)=
 \left(x_1,x_2,z+x_2,
 h+\frac{w_1}{T}x_1+w_2(a_\kappa x_2+b_\kappa z)\right)
\end{equation}
$T$ times. The only trainable tensor is $w=(w_1,w_2)\in\mathbb R^2$. At the start of occurrence $t$, $z=(t-1)x_2$, so its coefficients are
\[
 c_{t,1}=\frac1T,\qquad
 c_{t,2}=\frac1T+b_\kappa\left(t-\frac{T+1}{2}\right).
\]
Using $\sum_t(t-(T+1)/2)=0$ and
$\sum_t(t-(T+1)/2)^2=T(T^2-1)/12$ gives
\[
 \sum_t c_{t,j}=1,\qquad
 T\sum_t c_{t,1}^2=1,\qquad
 T\sum_t c_{t,2}^2=\kappa^2.
\]
Thus $f_w(x)=x^\top w$. The states $x_1,x_2,z$ have no parameter dependence, and every later $h$ update has derivative one with respect to its incoming $h$. Consequently the occurrence gradients satisfy $g_{t,j}=c_{t,j}g_j$.

Assign signal to the first coordinate and noise to the second. Take the design $\sqrt2e_1,\sqrt2e_2$ and responses $\sqrt2(\mu,\xi)$, where $\xi=\sigma\chi$ and $\chi$ is a symmetric sign. Population inputs are uniform on the design pair; their responses are respectively $\sqrt2\mu$ and $\sqrt2\sigma\chi'$, with an independent symmetric sign $\chi'$. With the empirical normalization $1/(2n)$ and $n=2$,
\begin{equation}\label{eq:occ-risk-target}
 \begin{aligned}
 F(w)&=\tfrac12\{(w_1-\mu)^2+(w_2-\xi)^2\},\\
 \mathcal R(w)&=\tfrac12\{(w_1-\mu)^2+w_2^2\},\\
 \tfrac12\mathbb E(f_w(X)-Y)^2&=\mathcal R(w)+\sigma^2/2.
 \end{aligned}
\end{equation}

\subsection{Proof of Theorem~\ref{thm:occ-first-risk}}
\label{sec:occurrence-first-proof}
Use the theorem's conditions $10^{-4}\le\mu\le.05$, $\sigma>0$, and $10^{-12}\le r=T^{-\beta}\le1$. Start at $w_0=0$ with zero moments and use the full training pair. Define
\[
 \begin{gathered}
 A=\frac{\mu}{\mu+\epsilon},\qquad
 B=\frac{\sigma}{\sigma+\epsilon},\qquad
 \tau=\frac{\sigma+\epsilon}{\kappa\sigma+\epsilon},\\
 J=A^2+B^2,\qquad H=A^2+\tau^2B^2,\qquad
 C=\mu A+\sigma\tau B,\qquad F_0=\frac{\mu^2+\sigma^2}{2},\\
 \gamma_*=\frac{\mu AC}{HF_0},\qquad
 \nu=B\sqrt{\frac{1-\tau^2}{J}},\qquad
 \Delta=\frac{\mu^2A^2B^2(1-\tau^2)}{2JH}>0.
 \end{gathered}
\]
The interval in Theorem~\ref{thm:occ-first-risk} is
\begin{equation}\label{eq:occ-risk-gamma}
 \gamma_*\left(1-\frac\nu2\right)
 \le\gamma\le
 \gamma_*\left(1+\frac\nu2\right).
\end{equation}
Since $0<\nu<1$, it is nonempty, positive, and independent of $\chi$. Every scale in this interval gives an accepted first trial and the pathwise margin
\begin{equation}\label{eq:occ-risk-gap}
 \mathcal R(w_O^+)
 \le\inf_{\eta\ge0}\mathcal R\{w_A^+(\eta)\}-\tfrac34\Delta
 =\frac{\mu^2B^2}{2J}-\tfrac34\Delta.
\end{equation}

\begin{proof}[Proof of Theorem~\ref{thm:occ-first-risk}]
At initialization, $g=(-\mu,-\chi\sigma)$. Bias correction cancels the zero-initialized moment coefficients, giving
\[
 \widehat m=g,\qquad
 \widehat v=(\mu^2,\sigma^2),\qquad
 \widehat q=(\mu^2,\kappa^2\sigma^2),\qquad
 a=-(A,\chi B),\quad u=-(A,\chi\tau B).
\]
Since $A\ge10000/10001>.9999$, $H\ge A^2>10^{-60}$ and the tiny-norm branch is inactive. Put $\rho=\sqrt{J/H}$. Norm matching gives
$d=-\rho(A,\chi\tau B)$ and $\|d\|=\sqrt J$. The group-scaled slope obeys
\[
 s=r\rho C\ge r\mu A
 \ge10^{-16}\frac{10000}{10001}>10^{-30},
\]
so neither fallback nor the slope floor is active.

Put $\lambda_*=\mu A/H$ and $\lambda_\gamma=\gamma F_0/C$. Equation~\eqref{eq:occ-risk-gamma} gives
$0<\lambda_\gamma\le(1+\nu/2)\lambda_*<3\lambda_*/2$.
The proposed loss-based step is $\alpha_{\rm loss}=\lambda_\gamma/(r\rho)$, and
\[
 \alpha_{\rm loss}r\|d\|
 =\lambda_\gamma\sqrt H
 <\frac{3\mu A}{2\sqrt H}
 \le\frac32\mu\le.075<.1.
\]
Because $w_0=0$ and $\|d\|\ge A>10^{-20}$, this proves $\alpha_{\rm loss}<c_B/r$. For the empty endpoint group, $c_E=.1/10^{-20}=10^{19}$, whereas
\[
 \alpha_{\rm loss}\le\frac{.075}{r\sqrt J}
 \le\frac{.075}{rA}
 <8\times10^{10}<c_E.
\]
Both movement caps are therefore inactive. The initial trial is
$w_O^+=\lambda_\gamma(A,\chi\tau B)$, and its training-loss decrease is
\[
 F_0-F(w_O^+)=\lambda_\gamma C-\tfrac12\lambda_\gamma^2H.
\]
Since $\lambda_\gamma H<1.5\mu A\le1.5C<1.9C$, this decrease exceeds
$.05\lambda_\gamma C=.05\alpha_{\rm loss}s$. The trial loss is finite, so the first of the eight allowed trials is accepted.

For every $\lambda\ge0$, completing the square gives
\[
 \mathcal R\{\lambda(A,\chi\tau B)\}
 =\frac{\mu^2\tau^2B^2}{2H}
  +\frac H2\left(\lambda-\frac{\mu A}{H}\right)^2.
\]
Adam's first output is $w_A^+(\eta)=\eta(A,\chi B)$; its minimum risk is
$\mu^2B^2/(2J)$, attained at $\eta=\mu A/J>0$. The difference between the two ray minima is
\[
 \frac{\mu^2B^2}{2J}-\frac{\mu^2\tau^2B^2}{2H}
 =\Delta.
\]
Moreover, $\lambda_\gamma/\lambda_*=\gamma/\gamma_*$, so the chosen interval implies
\[
 \frac H2(\lambda_\gamma-\lambda_*)^2
 =\frac{\mu^2A^2}{2H}\left(\frac\gamma{\gamma_*}-1\right)^2
 \le\frac{\mu^2A^2\nu^2}{8H}=\frac\Delta4.
\]
This proves \eqref{eq:occ-risk-gap}. The interval, margin, and Adam's minimizing step are independent of $\chi$, giving the claimed comparison of expected risks.
\end{proof}

For the example in Section~\ref{sec:occurrence-risk-main}, the endpoints of the exact interval \eqref{eq:occ-risk-gamma} are approximately $.0254$ and $.0530$. The choice $\gamma=1/32$ is therefore admissible before the noise sign is drawn.

\subsection{Risk Increase When Dispersion Attenuates Signal}
\label{v36:sec:signal-damping}
Keep block \eqref{eq:occ-risk-block} and its coefficients, but swap the signal and noise coordinates. Use training responses $\sqrt2(\xi,\mu)$; the population response at $\sqrt2e_1$ is now $\sqrt2\sigma\chi'$ and at $\sqrt2e_2$ is $\sqrt2\mu$. Then
\begin{equation}\label{v36:eq:swapped-risk-model}
 \begin{aligned}
 F_{\rm s}(w)&=\tfrac12\{(w_1-\xi)^2+(w_2-\mu)^2\},\\
 \mathcal R_{\rm s}(w)&=\tfrac12\{w_1^2+(w_2-\mu)^2\},\\
 \tfrac12\mathbb E(f_w(X)-Y)^2&=\mathcal R_{\rm s}(w)+\sigma^2/2.
 \end{aligned}
\end{equation}

\begin{proposition}[A reversal when dispersion attenuates signal]
\label{v36:prop:signal-damping}
Let $T\ge2$, $\kappa>1$, $\mu>0$, and $\sigma>0$ be finite. In \eqref{v36:eq:swapped-risk-model}, start at $w_0=0$ with zero moments, use the full training pair, and retain $\epsilon=10^{-8}$ and all safeguards of Algorithm~\ref{alg:occurrence-sgd}. The endpoint group is empty. For any finite $\gamma>0$ and $\beta\ge0$, put $r=T^{-\beta}>0$ and define
\[
 \begin{gathered}
 A=\frac{\mu}{\mu+\epsilon},\qquad
 B=\frac{\sigma}{\sigma+\epsilon},\qquad
 \tau_{\rm s}=\frac{\mu+\epsilon}{\kappa\mu+\epsilon},\\
 J=A^2+B^2,\qquad H_{\rm s}=B^2+\tau_{\rm s}^2A^2,\qquad
 \Delta_{\rm s}=\frac{\mu^2A^2B^2(1-\tau_{\rm s}^2)}{2JH_{\rm s}}>0.
 \end{gathered}
\]
For each $\chi\in\{-1,1\}$, every possible first LoopAda output, including the retained point after eight rejected trials, satisfies
\begin{equation}\label{v36:eq:signal-damping-risk}
 \mathcal R_{\rm s}(w_O^+)
 \ge\frac{\mu^2B^2}{2H_{\rm s}}
 =\inf_{\eta\ge0}\mathcal R_{\rm s}\{w_A^+(\eta)\}+\Delta_{\rm s},
\end{equation}
where $w_A^+(\eta)=\eta(\chi B,A)$. The strict comparison also holds after averaging over the training noise.
\end{proposition}

\begin{proof}
The first moments and directions are
\[
 \begin{gathered}
 g=\widehat m=(-\chi\sigma,-\mu),\qquad
 \widehat v=(\sigma^2,\mu^2),\qquad
 \widehat q=(\sigma^2,\kappa^2\mu^2),\\
 a=-(\chi B,A),\qquad u=-(\chi B,\tau_{\rm s}A).
 \end{gathered}
\]
Both normalization branches give $d=-\rho(\chi B,\tau_{\rm s}A)$, where
\[
 \rho=
 \begin{cases}
 \sqrt{J/H_{\rm s}},&H_{\rm s}>10^{-60},\\
 1,&H_{\rm s}\le10^{-60}.
 \end{cases}
\]
All of $A,B,\tau_{\rm s},\rho$ are positive and $\tau_{\rm s}<1$. The slope is
$s=r\rho(\sigma B+\mu\tau_{\rm s}A)>0$, so fallback is never used. The slope floor and both movement caps may be active and leave a finite positive initial scalar step $\alpha_0$: all three arguments of the minimum in Algorithm~\ref{alg:occurrence-sgd} are finite and positive in real arithmetic. At trial $\ell=0,\ldots,7$, the candidate is
\begin{equation}\label{v36:eq:swapped-trial-ray}
 w_\ell^+=\lambda_\ell(\chi B,\tau_{\rm s}A),
 \qquad \lambda_\ell=2^{-\ell}\alpha_0r\rho>0.
\end{equation}
Thus every accepted candidate lies on this ray, regardless of clipping, the slope floor, or backtracking. If all trials fail, the retained point is the same ray's origin, with $\lambda=0$. Retaining the new moments leaves this first output unchanged, so every branch of the first-update rule returns a point on the ray.

For every $\lambda\ge0$,
\begin{equation}\label{v36:eq:swapped-risk-square}
 \mathcal R_{\rm s}\{\lambda(\chi B,\tau_{\rm s}A)\}
 =\frac{\mu^2B^2}{2H_{\rm s}}
  +\frac{H_{\rm s}}2
   \left(\lambda-\frac{\mu\tau_{\rm s}A}{H_{\rm s}}\right)^2.
\end{equation}
Adam's ray has minimum $\mu^2B^2/(2J)$ at $\eta=\mu A/J>0$, and
\[
 \frac{\mu^2B^2}{2H_{\rm s}}-\frac{\mu^2B^2}{2J}
 =\frac{\mu^2A^2B^2(1-\tau_{\rm s}^2)}{2JH_{\rm s}}
 =\Delta_{\rm s}>0.
\]
This proves \eqref{v36:eq:signal-damping-risk} for every possible first output. The bound and Adam's minimizing step are independent of $\chi$, so the expected-risk comparison follows.
\end{proof}

\subsection{Gradient Dispersion at a Stationary Point}
\label{v36:sec:zero-aggregate}
\begin{example}[A zero aggregate gradient with available fitting parameters]
\label{v36:ex:zero-aggregate}
Apply the scalar affine block $h\mapsto\theta h+2$ twice, with fixed input $h_0=1$ and fixed readout $f=h_2$. Train only the shared scalar $\theta$ on one observation with response $y=2$ and half-squared loss. Then
\[
 f_L(\theta)=\theta^2+2\theta+2,\qquad
 F_L(\theta)=\tfrac12\{(\theta+1)^2-1\}^2.
\]
The looped model fits perfectly at $\theta=0$ and $\theta=-2$. At $\theta_0=-1$, the residual is $e=-1$. The first occurrence contributes $J_1=\theta_0h_0=-1$ to the prediction derivative, and the second contributes $J_2=h_1=1$. Thus
\[
 g_1=1,\quad g_2=-1,\quad g=0,\quad Q=4.
\]
With no endpoint parameters and zero initial moments, LoopAda has $m=v=0$ and $q>0$ after its first moment update, hence $a=u=d=0$. Its slope and fallback direction are zero, so no trial meets the positive-slope acceptance requirement. Repeating the same full batch leaves the aggregate gradient and first moment zero; the method stays at $\theta=-1$ at every iteration.

The example shows that disagreement can be positive at a stationary point even when another parameter value fits the sample exactly.
\end{example}

\subsection{Stalling under Finite Backtracking}\label{sec:supp-stall}
The eight-trial cap permits stalling at a nonstationary point when the minimum loss is positive. Consider one scalar shared tensor, two equal occurrence contributions, fixed endpoints, $r=1$, and $F(w)=1+(w-10^{-5})^2/2$. This is a smooth strongly convex squared-loss objective: two identical scalar design points with response coefficients $10^{-5}\pm\sqrt2$ give this normalization. Initialize $w=0$, all moments at zero, and $\gamma=1/32$. Equal contributions give $q=v$, so the proposed direction equals the internal Adam reference.

If an attempt is rejected, the parameter remains zero. Its gradient is therefore always $-10^{-5}$, and bias correction gives the same first and second moments at every attempt. The direction has magnitude $10^{-5}/(10^{-5}+10^{-8})$. The loss-based trial exceeds the movement cap, which binds at movement $.1$. The eight trial movements are $.1/2^j$, $j=0,\ldots,7$. For a positive movement $t$, the loss decrease is $10^{-5}t-t^2/2$; the acceptance test requires $t\le1.9\times10^{-5}$. Even the smallest trial, $.1/128$, exceeds this value. Every attempt is rejected, all subsequent parameters stay zero, and the point is nonstationary. The eight-trial rule therefore admits permanent stalling on a smooth strongly convex squared-loss objective with positive minimum.
\section{One-Update Prediction Risk}\label{app:general-risk}
Section~\ref{app:directional-risk} proves Theorem~\ref{thm:general-risk}.
The orthogonal-design comparison in Corollary~\ref{cor:general-family}
is proved in Sections~\ref{app:controller-risk-constants}
and~\ref{app:nonlinear-risk};
Section~\ref{app:dispersion-conditions} verifies the sufficient family
in~\eqref{gen:eq:family}.
\subsection{Proof of Theorem~\ref{thm:general-risk}}
\label{app:directional-risk}
We first give sufficient conditions for acceptance of the first trial,
then compare the expected risks by integrating their curvature bounds.

\paragraph{First-trial acceptance.}
At $w_0$ on sample $D$, write $g=\nabla F_D(w_0)$ for the
full-sample gradient and $g_t$ for its shared-block contribution
from application $t$. With zero initial moments, Adam's first
direction is $p_D=g/(|g|+\epsilon)$, with coordinatewise division.
For LoopAda, let $d_E$ and $d_B$ be the endpoint and shared-block
directions assigned tensorwise by Algorithm~\ref{alg:occurrence-sgd},
including its tiny-norm branch. For each shared tensor, the raw direction is
its aggregate gradient divided coordinatewise by
$\sqrt{T\sum_t g_t^{\odot2}}+\epsilon$, using its contributions
$g_t$. Put
\[
 d_D=(d_E,r d_B),\qquad s_D=g^\top d_D,\qquad
 v_D=F_D(w_0)d_D/s_D .
\]
Suppose $s_D>10^{-30}$ and
$\gamma F_D(w_0)/s_D<\min\{c_E,c_B/r\}$. The first trial
then returns $w_0-\gamma v_D$ if it satisfies sufficient decrease.
The empirical directional derivative at zero along this path is
$-g^\top v_D=-F_D(w_0)$. If the second derivative of
$t\mapsto F_D(w_0-tv_D)$ is bounded above by $H_D\ge0$ on
$[0,\Gamma]$, then
\[
 F_D(w_0)-F_D(w_0-\gamma v_D)
 \ge \gamma F_D(w_0)-\tfrac12 H_D\gamma^2 .
\]
Thus $\Gamma H_D\le 1.9 F_D(w_0)$ suffices for acceptance for all
$0<\gamma\le\Gamma$. The cap check at $\Gamma$ suffices throughout
this interval because the trial step is proportional to $\gamma$.

\Needspace{4\baselineskip}
\phantomsection\label{proof:theorem-one}
\begin{proof}[Proof of Theorem~\ref{thm:general-risk}]
Both expected risk curves start at $R(w_0)$. Integrating their
curvature bounds on the stated intervals gives
\begin{align*}
 R_A(\eta)&\ge R(w_0)-a_A\eta+\tfrac12 B_A\eta^2,\\
 R_O(\gamma)&\le R(w_0)-a_O\gamma+\tfrac12 B_O\gamma^2.
\end{align*}
The tail condition permits minimization of Adam's risk over
$[0,H]$ when $H$ is finite. For $H=\infty$, the quadratic bound
holds throughout $[0,\infty)$. In either case the first quadratic
attains its minimum at $t_A$, which gives \eqref{eq:directional-gap}.
Maximizing the right hand side over $0<\gamma\le\Gamma$ when $a_O>0$
gives the stated choice of scale.
\end{proof}

The directional curvature has a prediction-space interpretation.
When the derivatives can be passed through the integral, for a
direction $d$ it is
\[
 D_d^2 R(w)=\E_X\!\left[(D_d f_w(X))^2
                      +(f_w(X)-f_0(X))D_d^2 f_w(X)\right].
\]
The first term measures the size of changes in prediction. The
second measures how the prediction error interacts with curvature
of the fitted model. Both are evaluated against the true regression
function. All rates in the theorem are fixed before drawing $D$.

\paragraph{Comparison relative to an approximant.}
Subtracting the approximation error of the common fitted family from
both risks preserves their difference. An approximation bound also gives
a direct risk comparison. Let $f^\dagger$ be a fixed
looped approximant with $\|f^\dagger-f_0\|_{L^2(P_X)}\le\delta$.
If the predictors trained on the actual responses have root mean
squared distances to $f^\dagger$ bounded above by $u$ for LoopAda and
below by $\ell$ for all Adam rates, then the root mean squared
distances to $f_0$ are at most $u+\delta$ and at least
$(\ell-\delta)_+$, respectively. These bounds follow from the
triangle and reverse triangle inequalities in $L^2(P_D\times P_X)$.
The comparison therefore transfers whenever $\ell-u>2\delta$.

\subsection{A Bounded Nonlinear Predictor}
\label{app:bounded-loop-example}
Use the orthogonal design and the shared block in \eqref{gen:eq:block}
with two predictors, dispersion factors $1$ and $10$, and fixed
endpoints. On $x_j=\sqrt{2}e_j$ the scalar output of the block after
$T$ applications is $h_T=\sqrt{2}w_j$. Use the fixed readout
$h_T\mapsto\sqrt{2}\psi(h_T/\sqrt{2})$, where
$\psi(s)=2b\tanh(s/(2b))$ and $b=.01$.
The normalized training responses are $(b,\chi b)$ for a fair sign
$\chi$, and the normalized regression means are $(b,0)$. At zero
weights, $\psi'(0)=1$, so the occurrence gradients retain the
fractions $c_{t,j}$ in the linear model.
Set $r=1$, $\epsilon=10^{-8}$, and write
\[
 p=\frac{b}{b+\epsilon},\qquad h=\frac{b}{10b+\epsilon},\qquad
 q=\frac{p}{p+h}.
\]
Thus $.908<q<.91$. The first Adam output is $(s,\chi s)$ with
$s=p\eta$. Its risk is $\{(\psi(s)-b)^2+\psi(s)^2\}/2$.
It is nondecreasing for $s\ge s_*=2b\operatorname{arctanh}(1/4)$
because its derivative in $s$ is $(2\psi(s)-b)\psi'(s)$.
On $[0,s_*]$, its second derivative is at least
$2(15/16)^2$ since $\psi'\ge15/16$ and
$(2\psi-b)\psi''\ge0$. In Theorem~\ref{thm:general-risk} we may
take $H=s_*/p$, $B_A=2(15/16)^2p^2$, and $a_A=bp$.
The risk on the Adam path is bounded, and its best possible
improvement is at most $a_A^2/(2B_A)=64b^2/225$ by this bound.

The proposed first LoopAda output is
$\gamma b(q,\chi(1-q))$ by the loss-based step-size rule.
The pooled direction has squared norm at least $p^2>.99$.
Its norm-matched descent slope exceeds $.01$ since norm matching
scales the pooled direction by at least one and $b(p+h)>.01$.
For $0<\gamma\le1$, the trial displacement is at most $b<.1$
and its step size is below $.01$, so both caps and the slope floor
are inactive. Write $z=\tanh(s/(2b))$ for $s\ge0$. The second
derivatives of $(\psi(s)-b)^2/2$ and $\psi(s)^2/2$ are
$(1-z^2)(1+z-3z^2)\le13/12$ and
$(1-z^2)(1-3z^2)\le1$, respectively.
The first bound follows by completing the square:
$1+z-3z^2=13/12-3(z-1/6)^2$.
These inequalities give empirical directional curvature at most
$(13/12)b^2$ along the trial path. Since the initial loss is
$b^2$, the first trial satisfies the acceptance criterion for every
$0<\gamma\le1$.

For prediction risk, take $a_O=qb^2$,
$B_O=b^2\{(13/12)q^2+(1-q)^2\}$, and $\Gamma=1$.
At $\gamma=1$ the LoopAda risk improvement is at least
$a_O-B_O/2>.45b^2$ since $.908<q<.91$. Subtracting the bound
$64b^2/225$ on Adam's best improvement in
Theorem~\ref{thm:general-risk} gives a risk gap greater than
$.165b^2$ in favor of LoopAda, over all fixed Adam rates.

\subsection{Nonlinear Prediction with Approximation Error}
\label{app:holder-loop-example}
Consider orthogonal features. Let
$X_1,X_2,X_3$ be independent fair signs, put $U=X_1X_2$ and
$V=X_2X_3$, and let $Z$ be independently uniform on $[0,1]$.
The features $U,V,UV$ have mean zero and are orthonormal. Let
$b=.01$ and $f_0(X,Z)=bU+h(Z)$, where $h$ is a centered
H\"older-smooth function with $\|h\|_\infty\le .005b$.
This regression function is H\"older smooth on the ambient compact
domain for any order admitted by $h$. We use $n=31$ independent
observations and independent fair response noise $\varepsilon=\pm .005b$.

The fitted family is
\[
 f_w(X,Z)=\psi(w_1)U+\psi(w_2)V+(w_1w_2/b)UV,\qquad
 \psi(s)=s+4s^3/b^2 .
\]
It is realized by two applications of one shared block with trainable
vector $w$. Initialize the state as $(U,V,0,0)$ and copy $U,V$
unchanged at each application. Update the vector $z$ and scalar $q$
in the remaining state by
\[
 (z,q)\longmapsto
 \bigl(Az+Bw,\;(B^{-1}z)_1(B^{-1}z)_2/b\bigr),
 \quad A=\operatorname{diag}(1,-9/11),\quad B=(I+A)^{-1}.
\]
At the second application, $z=w$ and $q=w_1w_2/b$. Use the fixed
readout $\psi(z_1)U+\psi(z_2)V+qUV$. At zero weights the
interaction has derivative zero and $\psi'(0)=1$. The first and
second applications therefore contribute fractions $(1/2,1/2)$
of the aggregate gradient in coordinate 1 and $(-9/2,11/2)$
in coordinate 2. This gives dispersion factors $1$ and $\sqrt{101}$.

Orthogonality gives the population risk
\[
 R(w)=\tfrac12\|h\|_2^2+\tfrac12\bigl\{
          (\psi(w_1)-b)^2+\psi(w_2)^2+(w_1w_2/b)^2\bigr\}.
\]
The approximation error is exactly $\|h\|_2^2/2$, attained at
$w=(b/2,0)$. At this weight the cubic term supplies half of the
fitted signal $bU$.

\paragraph{Adam's risk at every rate.}
Define $a_i=Y_iU_i/b\in[.99,1.01]$ and write
$H=n^{-1}\sum_i a_i$, $J=n^{-1}\sum_i a_i U_iV_i$, and
$K=n^{-1}\sum_i a_i^2$. Then $g=-b(H,J)$ at zero. The odd
sample size gives $|J|\ge 1/31-.01$. Also $|J|\le H$ and
$K/H\in[.99,1.01]$. For $\chi=\operatorname{sign}(J)$ and
$\delta=\epsilon/b=10^{-6}$, the Adam first output can be written
as $(s,\chi\nu s)$, where $s\ge0$ and
\[
 \nu=\frac{1+\delta/H}{1+\delta/|J|}\in(1-5\times10^{-5},1].
\]
Since $\psi$ is odd and $\psi(\nu s)\ge\nu^3\psi(s)$ for
$s\ge0$, minimizing a quadratic in $\psi(s)$ gives
\[
 R(s,\chi\nu s)-\tfrac12\|h\|_2^2
 \ge \frac{\nu^6b^2}{2(1+\nu^6)}>.2499b^2 .
\]

\paragraph{LoopAda's first trial.}
Take $r=1$ and use the denominator constant $\epsilon=10^{-8}$.
Write $\rho=|J|/H$ and
\[
 \tau=\frac{1+\delta/H}{\sqrt{101}+\delta/|J|}<.101 .
\]
The unclipped trial from zero has weights $(bv,\chi b\tau v)$ with
$v=\gamma(K/H)/\{2(1+\rho\tau)\}$. For every
$\gamma\in[1.01,1.08]$, the bounds on $K/H$, $\rho$ and $\tau$
give $.45<v<.55$. The squared norm of the raw pooled direction
exceeds $.99$ because its first coordinate has magnitude
$H/(H+\delta)$. The nonnegative norm-matching rescaling is at least
one. The matched slope therefore exceeds $.0098$, and
$F_D(0)=b^2K/2$ gives a loss-based step size below $.0057$.
The block cap exceeds $.0707$ since the matched direction has
norm below $\sqrt2$. The endpoint cap is $10^{19}$.

At the proposed weights, the three errors in the orthogonal features are,
up to signs, $b(v+4v^3-1)$, $b\{\tau v+4(\tau v)^3\}$, and
$b\tau v^2$. On $.45<v<.55$ and $0<\tau<.101$, their magnitudes
are bounded by $.216b$, $.057b$, and $.031b$, respectively.
Since $|Y_i-bU_i|\le.01b$, the trial-to-initial
training-loss ratio is at most $(.314/.99)^2<.101$. This satisfies
the acceptance rule, which allows a ratio of $1-.05\gamma$.
The resulting population risk satisfies
\[
 R(w_O^+)-\tfrac12\|h\|_2^2
 \le \tfrac12(.216^2+.057^2+.031^2)b^2<.026b^2 .
\]
Thus the same LoopAda scale interval gives a lower risk for every
realized sample and every Adam rate in this family. The common
approximation error cancels in the comparison.

\subsection{Linear Model with Noisy Observations}

Partition $\{1,\ldots,d\}=S\mathbin{\dot\cup}N$, with $d_S=|S|\ge1$ and $d_N=|N|\ge1$. Let $w^\star_j\ne0$ for $j\in S$ and $w^\star_j=0$ for $j\in N$. Write $b_j=|w^\star_j|$. Let all signs $\xi_j$ be independent, equiprobable $\{-1,1\}$ variables and define the training coefficient vector
\begin{equation}
 Y_j=w^\star_j+\delta_j\xi_j\quad(j\in S),\qquad
 Y_j=\sigma_j\xi_j\quad(j\in N),
\end{equation}
where $\delta_j\ge0$ and $\sigma_j>0$. The training sample consists of inputs $x_j=\sqrt d\,e_j$ and responses $\sqrt d\,Y_j$. A linear predictor $f_w(x)=x^\top w$ therefore has empirical half-squared loss
\begin{equation}
 F_Y(w)=\frac1{2d}\sum_{j=1}^d\{f_w(x_j)-\sqrt d\,Y_j\}^2
       =\frac12\|w-Y\|^2.
\end{equation}
Population inputs are uniform on the same design, with independent response noise of the same coordinatewise law. The excess population half-squared risk is
\begin{equation}
 \mathcal R(w)=\frac12\|w-w^\star\|^2.
\end{equation}
The population loss adds the constant $(\sum_{S}\delta_j^2+\sum_N\sigma_j^2)/2$ to \eqref{gen:eq:risk}.

The predictor is realized by one identical block used $T\ge2$ times. Choose dispersion factors $\kappa_j\ge1$, put
\[
 B_j=\sqrt{\frac{12(\kappa_j^2-1)}{T^2(T^2-1)}},\qquad
 A_j=\frac1T-\frac{T-1}{2}B_j,
\]
and use the fixed encoder $x\mapsto(x,0,0)$, fixed readout $(x,z,h)\mapsto h$, and shared block
\begin{equation}\label{gen:eq:block}
 F_w(x,z,h)=\left(x,z+x,h+\sum_{j=1}^d w_j(A_jx_j+B_jz_j)\right).
\end{equation}
The only trainable tensor is the entire vector $w\in\R^d$. At occurrence $t$, $z=(t-1)x$, giving coefficients
\[
 c_{tj}=\frac1T+B_j\left(t-\frac{T+1}{2}\right),\qquad
 \sum_t c_{tj}=1,\qquad T\sum_t c_{tj}^2=\kappa_j^2.
\]
Thus $f_w(x)=x^\top w$ and, at every parameter point, the occurrence gradient is $g_{t,j}=c_{tj}g_j$. Each block is linear in its hidden state for fixed $w$. One shared tensor and fixed endpoints define the parameterization.

\subsection{Proof of Corollary~\ref{cor:general-family}: Linear Case}
\label{app:controller-risk-constants}

The proof identifies the accepted first update, evaluates its exact
expected risk, and derives admissible scales with a positive risk gap.

\paragraph{Notation and preliminary bounds.}
The following bounds give a sufficient parameter range for the unclipped
first-trial analysis.
\begin{equation}
 \begin{gathered}
 0\le\delta_j\le b_j/10,\quad b_j-\delta_j\ge10^{-4}\quad(j\in S),
 \|w^\star\|\le .05,\qquad 1\le\kappa_j\le10\quad(j\in S),\\
 \kappa_j\ge1,\ \sigma_j>0\quad(j\in N),\qquad
 10^{-12}\le r=T^{-\beta}\le1,\qquad \epsilon=10^{-8}.
 \end{gathered}
\end{equation}
The number of zero coefficients is unrestricted. The relative-noise bound
preserves the signal signs, while the lower amplitude bound gives
$\epsilon/(b_j-\delta_j)\le10^{-4}$ and controls the influence of the
denominator constant. The norm, dispersion, and block-scaling bounds
supply uniform margins for the displacement and descent-slope inequalities
derived below. Those inequalities also provide sufficient bounds for other
amplitude and dispersion ranges.

For $j\in S$, define $b_j^\pm=b_j\pm\delta_j$ and
\[
 p_j^\pm=\frac{b_j^\pm}{b_j^\pm+\epsilon},\qquad
 h_j^\pm=\frac{b_j^\pm}{\kappa_jb_j^\pm+\epsilon}.
\]
For $j\in N$, put $p_j=\sigma_j/(\sigma_j+\epsilon)$ and
$h_j=\sigma_j/(\kappa_j\sigma_j+\epsilon)$. Introduce four finite, closed-form sums:
\begin{equation}\label{gen:eq:moments}
 \begin{aligned}
 m_A&=\sum_{j\in S} b_j\frac{p_j^-+p_j^+}{2},&
 v_A&=\sum_{j\in S}\frac{(p_j^-)^2+(p_j^+)^2}{2}+\sum_{j\in N}p_j^2,\\
 m_O&=\sum_{j\in S} b_j\frac{h_j^-+h_j^+}{2},&
 v_O&=\sum_{j\in S}\frac{(h_j^-)^2+(h_j^+)^2}{2}+\sum_{j\in N}h_j^2.
 \end{aligned}
\end{equation}
The controller bounds are
\begin{equation}
 \begin{aligned}
 L_\pm&=\tfrac12\left\{\sum_{j\in S}(b_j^\pm)^2+\sum_{j\in N}\sigma_j^2\right\},\\
 C_\pm&=\sum_{j\in S}b_j^\pm h_j^\pm+\sum_{j\in N}\sigma_jh_j,\\
 a_-&=L_-/C_+,\qquad a_+=L_+/C_-,\qquad \rho=a_-/a_+.
 \end{aligned}
\end{equation}
These constants are determined by amplitudes, noise levels, and dispersion factors before the training signs are drawn. Set
\begin{equation}\label{gen:eq:conservative-scale}
 G_A=\frac{m_A^2}{v_A},\qquad
 G_O=\rho^2\frac{m_O^2}{v_O},\qquad
 \gamma_c=\frac{a_-m_O}{a_+^2v_O}.
\end{equation}
The attenuation factor $\rho$ converts the comparison of optimized rays into a bound for the loss-based update by accounting for variation in its scalar multiplier.

The conservative scale interval and gap used below when $G_O>G_A$ are
\[
 \tau_\gamma=\tfrac12\sqrt{1-G_A/G_O},\qquad
 I_0=[\gamma_c(1-\tau_\gamma),\gamma_c(1+\tau_\gamma)],\qquad
 M_0=\tfrac38(G_O-G_A).
\]

\Needspace{5\baselineskip}
\phantomsection\label{proof:corollary-linear}
\begin{proof}
\textit{Step 1: First-update directions and controller bounds.}
For a realized $Y$, define signed vectors
\[
 p_j(Y)=\frac{Y_j}{|Y_j|+\epsilon},\qquad
 h_j(Y)=\frac{Y_j}{\kappa_j|Y_j|+\epsilon},\qquad
 H(Y)=\|h(Y)\|^2,
\]
and $C(Y)=Y^\top h(Y)>0$, $L(Y)=\|Y\|^2/2$, $a(Y)=L(Y)/C(Y)$. All signal signs are stable under \eqref{gen:eq:safe}. The increasing functions $x\mapsto x/(\kappa x+\epsilon)$ and $x\mapsto x^2/(\kappa x+\epsilon)$ yield
\begin{equation}\label{gen:eq:pointwise-bounds}
 L_-\le L(Y)\le L_+,\quad C_-\le C(Y)\le C_+,\quad
 a_-\le a(Y)\le a_+.
\end{equation}
Moreover,
\[
 (w^\star)^\top\E p(Y)=m_A,\quad \E\|p(Y)\|^2=v_A,
 \quad (w^\star)^\top\E h(Y)=m_O,\quad \E H(Y)=v_O.
\]
In particular, $(w^\star)^\top h(Y)>0$ for every realization.

At the first update the bias corrections cancel the $.9/.999$ coefficients:
\[
 \widehat m=-Y,\qquad \widehat v_j=Y_j^2,\qquad
 \widehat q_j=\kappa_j^2Y_j^2.
\]
Thus the internal Adam reference is $-p(Y)$, the raw occurrence direction is $-h(Y)$, and the norm-matched direction is
\[
 d=-R(Y)h(Y),\qquad R(Y)=\|p(Y)\|/\|h(Y)\|\ge1.
\]
The relative block factor gives slope $s_0=rR(Y)C(Y)>0$.

\medskip\noindent
\textit{Step 2: First-trial acceptance.}
For the safeguards, put
\[
 m=\min_{j\in S}b_j^-,\quad K_s=\max_{j\in S}\kappa_j,\quad
 \omega=1+\epsilon/m\le1.0001,
\]
and
\[
 H_-=\sum_{S}(h_j^-)^2+\sum_Nh_j^2,\qquad
 H_+=\sum_{S}(h_j^+)^2+\sum_Nh_j^2.
\]
Then
\begin{equation}\label{gen:eq:guard-elements}
 \begin{gathered}
 H_-\ge\frac{d_S}{(K_s+\epsilon/m)^2}\ge\frac1{10.0001^2}>10^{-60},\quad
 \|p(Y)\|^2\ge\frac{d_S}{(1+\epsilon/m)^2}>10^{-40},\\
 C_-\ge\frac{d_S m}{K_s+\epsilon/m},\qquad
 s_0\ge rC_->9\times10^{-18}>10^{-30}.
 \end{gathered}
\end{equation}
Consequently the tiny-norm branch, the nonpositive-slope fallback, the cap's tiny-direction denominator, and the slope floor are inactive.

For each signal coordinate, $h_j^+/h_j^-\le\omega$, since
$h_j^-\ge1/(\kappa_j+\epsilon/m)$ and $h_j^+\le1/\kappa_j$.
It follows that
\begin{equation}\label{gen:eq:ratios}
 H_+/H_-\le\omega^2,\qquad
 m_O/C_-\le\omega/(1-.1).
\end{equation}
For the second inequality, use
$b_j\le b_j^-/(1-.1)$ and $\E|h_j(Y)|\le h_j^+\le\omega h_j^-$,
then add the nonnegative nuisance contribution to $C_-$. Also Cauchy--Schwarz gives
\begin{equation}\label{gen:eq:cs}
 m_O\le\|w^\star\|\sqrt{v_O},\qquad
 H_-\le v_O\le H_+.
\end{equation}

For each noise realization, let $t=\gamma a(Y)$. The proposed
displacement is $t h(Y)$, and its loss-based step size is
$\gamma a(Y)/(rR(Y))$. The initial block movement cap is $.1$ in
parameter space since $\|d\|=\|p\|>10^{-20}$. The empty endpoint
group has cap $10^{19}$. Because
$F_Y(0)-F_Y(th)=tC(Y)-t^2H(Y)/2$, strict sufficient decrease holds
if $tH(Y)/C(Y)<1.9$, where $1.9=2(1-.05)$ follows from the
Armijo coefficient $.05$. Thus the following bound incorporates both
caps and the loss check:
\begin{equation}\label{gen:eq:safe-scale}
 \Gamma_{\rm safe}=\min_Y\left\{
 \frac{.1}{a(Y)\sqrt{H(Y)}},
 \frac{10^{19}rR(Y)}{a(Y)},
 \frac{1.9 C(Y)}{a(Y)H(Y)}
 \right\}.
\end{equation}
The minimum is over all noise realizations. For any
$0<\gamma<\Gamma_{\rm safe}$ the first trial is accepted without
clipping or fallback and
\begin{equation}\label{gen:eq:exact-output}
 w_O^+(\gamma)=\gamma a(Y)h(Y).
\end{equation}
\medskip\noindent
\textit{Step 3: Exact expected-risk comparison.}
Adam's first output is $\eta p(Y)$. Minimizing its expected
quadratic risk over $\eta\ge0$ gives
$\|w^\star\|^2/2-G_A/2$, since $(w^\star)^\top\E p(Y)>0$.
Expanding LoopAda's quadratic risk at \eqref{gen:eq:exact-output}
now gives exactly the gap $M(\gamma)$.

\medskip\noindent
\textit{Step 4: Admissible scales and positive margins.}
The maximizer $\gamma_*=\widetilde m/\widetilde v$ of $M(\gamma)$
is strictly within the safeguarded range. Indeed, put
$W(Y)=(w^\star)^\top h(Y)>0$. The coordinate amplitude bounds give
$m_O/\omega\le W(Y)\le\omega m_O$ and
$v_O/\omega^2\le H(Y)\le\omega^2v_O$. Consequently,
\[
 \gamma_* a_+\le\omega^3\frac{m_O}{v_O}\frac{a_+}{\E a(Y)}
 \le\omega^3\frac{m_O}{v_O}\frac{L_+}{\E L(Y)}\frac{C_+}{C_-}.
\]
The first inequality uses $\E a(Y)^2\ge(\E a(Y))^2$.
The second uses $a(Y)\ge L(Y)/C_+$.
The signal-noise bound $\delta_j\le .1b_j$ gives
$L_+/\E L(Y)\le1.21$ and $C_+/C_-\le(11/9)\omega$.
As $\omega\le1.0001$, this shows
$\gamma_*a_+<1.49m_O/v_O$. At any scale with
$\gamma a_+\le1.5m_O/v_O$, \eqref{gen:eq:ratios} and
\eqref{gen:eq:cs} give
\[
 \begin{gathered}
 t\sqrt{H(Y)}\le1.5\omega\|w^\star\|<.1,\qquad
 \frac{\gamma a(Y)}{rR(Y)}\le
 \frac{1.5\|w^\star\|}{r\sqrt{H_-}}<10^{19},\\
 \frac{tH(Y)}{C(Y)}\le\frac{1.5\omega^3}{.9}<1.9 .
 \end{gathered}
\]
Thus $\gamma_*<\Gamma_{\rm safe}$. The same inequalities hold on
$I_0$ when $G_O>G_A$, since
$\gamma a_+\le1.5\rho m_O/v_O\le1.5m_O/v_O$ there.

For the conservative bound, expand the risk at
\eqref{gen:eq:exact-output} and use $a_-\le a(Y)\le a_+$:
\[
 \E\mathcal R(w_O^+(\gamma))\le
 \tfrac12\|w^\star\|^2-\tfrac12G_O
       +\tfrac12a_+^2v_O(\gamma-\gamma_c)^2.
\]
The final term is at most $(G_O-G_A)/8$ on $I_0$.
Thus $M(\gamma)\ge M_0$ throughout $I_0$.
\end{proof}

\subsection{Proof of Corollary~\ref{cor:general-family}: Nonlinear Case}
\label{app:nonlinear-risk}
In the block \eqref{gen:eq:block}, replace the increment
$u=\sum_j w_j(A_jx_j+B_jz_j)$ by $u+\lambda\varphi(u)$. The other
state components still carry $x$ and its accumulated copies $z$.
Thus on $x_j=\sqrt d\,e_j$, application $t$ uses
$u=\sqrt d\,c_{t,j}w_j$. Summing and dividing by $\sqrt d$ gives
\eqref{gen:eq:response-link}. The encoder and readout are fixed and $w$
remains one trainable tensor.

\Needspace{5\baselineskip}
\phantomsection\label{proof:corollary-nonlinear}
\begin{proof}
\textit{Step 1: Persistence of first-trial acceptance.}
The normalized training loss is
$F_\lambda(w)=\|\psi_\lambda(w)-Y_\lambda\|^2/2$.
At $w=0$, the derivative of the increment with respect to $u$ is one,
since $\varphi'(0)=0$. The derivative of the normalized prediction on
$x_j$ from application $t$ is therefore $c_{t,j}$. Since
$\psi_{\lambda,j}(0)=0$, the occurrence contribution to coordinate $j$
of the training gradient is $-c_{t,j}Y_{\lambda,j}$. It follows that
Adam's first output is $\eta p(Y_\lambda)$ and LoopAda's norm-matched
direction is $-D_\lambda h(Y_\lambda)$, where
$D_\lambda=\|p(Y_\lambda)\|/\|h(Y_\lambda)\|$. With the slope floor
and movement caps inactive, the trial is
\begin{equation}\label{gen:eq:nonlinear-trial}
 u_{\lambda,\gamma}
 =\gamma\,\frac{\|Y_\lambda\|^2}
                 {2Y_\lambda^\top h(Y_\lambda)}\,h(Y_\lambda).
\end{equation}
The cap for the empty endpoint group is $10^{19}$ as in the linear
proof. On the closed interval $I$ in Corollary~\ref{cor:general-family}, the first-trial inequalities are strict at $\lambda=0$,
including the norm and slope thresholds, the two cap comparisons,
and sufficient decrease. The proposed displacement, the trial loss,
and these inequalities depend continuously on $(\lambda,\gamma)$
near $\{0\}\times I$ for each sign pattern. Their denominators are
positive there by the linear proof. The sign set is finite and
$I$ is compact. Thus all these inequalities continue to hold
uniformly for sufficiently small $\lambda\ge0$. In this range
$w_O^+(\gamma,\lambda)=u_{\lambda,\gamma}$.

\medskip\noindent
\textit{Step 2: A lower risk bound over all Adam rates.}
Write $\ell_\lambda=1-\lambda C_\varphi$ and take $\lambda$ small
enough that $\ell_\lambda>0$. By \eqref{gen:eq:nonlinear-slope},
$|\psi_{\lambda,j}(s)-\psi_{\lambda,j}(t)|
\ge \ell_\lambda |s-t|$ for all real $s,t$. Substituting
Adam's first output and taking the infimum gives
\begin{align}
 \inf_{\eta\ge0}\E\mathcal R_\lambda\{\eta p(Y_\lambda)\}
 &\ge \ell_\lambda^2 B_\lambda,\label{gen:eq:nonlinear-adam}\\
 B_\lambda
 &=\tfrac12\|w^\star\|^2
   -\frac{\{(w^\star)^\top\E p(Y_\lambda)\}_+^2}
          {2\E\|p(Y_\lambda)\|^2},\notag
\end{align}
where $\{a\}_+=\max\{a,0\}$ denotes the positive part.
The bound follows by minimizing the quadratic
$\E\|\eta p(Y_\lambda)-w^\star\|^2/2$ over $\eta\ge0$.
Its denominator is positive at zero, and $B_0$ is the exact
optimal linear Adam risk.

\medskip\noindent
\textit{Step 3: Preservation of the uniform risk gap.}
The finite sign support gives continuity
of $\ell_\lambda^2B_\lambda$ at zero and continuity of the expected
risk at \eqref{gen:eq:nonlinear-trial}, uniformly over $\gamma\in I$.
The linear gap is at least $M_I$ on $I$. Choose $\lambda_0>0$ small
enough that the changes in the Adam lower bound and in the LoopAda
risk are each at most $M_I/4$, while retaining the strict first-trial
inequalities. This gives \eqref{gen:eq:main-bound} throughout
$[0,\lambda_0]\times I$.
\end{proof}

\paragraph{Examples of the slope condition.}
If $|\varphi'|\le L$, then \eqref{gen:eq:nonlinear-slope} holds with
$C_\varphi=L\max_j\sum_t|c_{t,j}|$. In particular, taking
$\varphi(u)=\tanh(u)-u$ gives the increment
$(1-\lambda)u+\lambda\tanh(u)$. To verify the odd-power case,
pair the coefficients symmetric about $1/T$ in \eqref{gen:eq:block}.
In each pair, two nonnegative coefficients give a nonnegative sum
of odd powers. If one coefficient is negative, the positive
coefficient is larger in magnitude, and the sum of odd powers is
positive. An unpaired central coefficient is $1/T$. Thus
$\sum_t c_{t,j}^{\,m}>0$ for odd $m\ge3$. For
$\varphi(u)=a u^m$ with $a>0$, the derivative of
$\psi_{\lambda,j}(s)-s$ is
$\lambda a m d^{(m-1)/2}s^{m-1}\sum_t c_{t,j}^{\,m}\ge0$.

\paragraph{A finite cubic example.}
Take $d=3$, $T=8$, $r=1$, $\kappa=(1,8,8)$, and
$w^\star=(.01,0,0)$. The signal coordinate has zero noise and each
nuisance coordinate has independent sign noise of amplitude $.1$.
For $\varphi(u)=u^3$, symmetry about $1/T$ gives
$\sum_t c_{t,j}^3=(3\kappa_j^2-2)/T^2$. Set $\lambda=10^5$ and
$\gamma=1/32$. The normalized signal mean is $47/3200$,
approximately $46.9\%$ above its linear value. Rational evaluation of each
of the four noise patterns gives first-trial acceptance without clipping
or fallback. The expected excess LoopAda risk is approximately
$1.303\times10^{-5}$. Equation~\eqref{gen:eq:nonlinear-adam} bounds the
optimal fixed-rate Adam risk strictly above $3.333\times10^{-5}$.

\subsection{Sufficient Conditions on Gradient Dispersion}
\label{app:dispersion-conditions}

The next conditions relate the gain to the dispersion of gradient
contributions, the numbers of signal and nuisance predictors, and
variation in the step-size multiplier.

\begin{corollary}[Dispersion contrast, coordinate counts, and controller variability]\label{gen:cor:contrast}
Under \eqref{gen:eq:safe}, let
\[
 k_s=\min_{j\in S}\kappa_j,\quad K_s=\max_{j\in S}\kappa_j,
 \quad K_n=\min_{j\in N}\kappa_j,\quad \sigma_* =\min_{j\in N}\sigma_j,
\]
and define
\[
 P=\frac{d_S}{(1+\epsilon/m)^2}+\frac{d_N}{(1+\epsilon/\sigma_*)^2},\qquad
 D=(K_s+\epsilon/m)^2\left(\frac{d_S}{k_s^2}+\frac{d_N}{K_n^2}\right).
\]
If
\begin{equation}\label{gen:eq:alignment}
 \boxed{\rho^2 P>D,}
\end{equation}
then the linear risk margin on $I_0$ is at least
\begin{equation}\label{gen:eq:coarse-margin}
 \frac38\|w^\star\|_1^2\left(\frac{\rho^2}{D}-\frac1P\right)>0.
\end{equation}
Under \eqref{gen:eq:nonlinear-slope}, the nonlinear conclusion of
Corollary~\ref{cor:general-family} also holds on $I_0$.
\end{corollary}
\begin{proof}
The coordinate formulas give
\[
 m_A\le\|w^\star\|_1,\quad v_A\ge P,\quad
 m_O\ge\frac{\|w^\star\|_1}{K_s+\epsilon/m},\quad
 v_O\le\frac{d_S}{k_s^2}+\frac{d_N}{K_n^2}.
\]
Consequently $G_A\le\|w^\star\|_1^2/P$ and
$G_O\ge\rho^2\|w^\star\|_1^2/D$. These imply both claims.
\end{proof}

\paragraph{Verification of the sufficient family in~\eqref{gen:eq:family}.}
\label{proof:sufficient-family}
In \eqref{gen:eq:family}, the dispersion and count bounds control the
contribution of nuisance coordinates to the update, while the noise
amplitudes control variation in the common multiplier $a(Y)$. Together
they imply $\widetilde m^2/\widetilde v>G_A$ and yield an interval of
accepted scales with positive risk gaps. The proof reduces this comparison
to an affine inequality in the ratio of nuisance to signal coordinates.
\begin{proof}
Put $B_S=\sum_S b_j^2$, $B_N=\sum_N\sigma_j^2$,
$x=d_N/d_S\ge1/2$, and $\omega_0=1.0001$.
Then $B_N\ge x B_S$. Under the $5\%$ signal-noise bound,
the lower and upper bounds for $L$ and $C$ imply
\[
 \rho\ge\rho_0=
 \frac{(19/20)^2+1/2}{(21/20)^2+1/2}\frac{19}{21\omega_0}.
\]
For $L_-/L_+$ this follows by bounding $B_N/B_S$ below by
$1/2$. For $C_-/C_+$ each signal term is at least
$19/(21\omega_0)$ times its upper bound, and the nuisance terms coincide.

Since $a(Y)\in[a_-,a_+]$, the inequality
$\E a(Y)^2\le(a_-+a_+)\E a(Y)-a_-a_+$ gives
\[
 \frac{(\E a(Y))^2}{\E a(Y)^2}\ge
 \frac{4a_-a_+}{(a_-+a_+)^2}=\frac{4\rho}{(1+\rho)^2}.
\]
Combining this with $W(Y)\ge m_O/\omega$, $H(Y)\le\omega^2v_O$,
and the coordinate bounds for $P,D$ gives
\[
 \frac{\widetilde m^2}{\widetilde v}\ge
 \frac{4\rho_0}{(1+\rho_0)^2\omega_0^4}
          \frac{\|w^\star\|_1^2}{D},
 \qquad G_A\le\frac{\|w^\star\|_1^2}{P}.
\]
Here $P\ge(d_S+d_N)/\omega_0^2$ and
$D\le 1.2001^2(d_S+d_N/36)$. To compare the gains, set
$A=4\rho_0/\{(1+\rho_0)^2\omega_0^6\}$ and $B=1.2001^2$.
The difference $A(1+x)-B(1+x/36)$ exceeds $.018$ at $x=1/2$
and has slope greater than $.94$, by exact rational arithmetic.
Hence the exact gain $\widetilde m^2/\widetilde v$ exceeds $G_A$.

For this family the earlier bound for $\gamma_*a_+$ improves to
\[
 \gamma_*a_+\le
 \omega_0^4\frac{1.05^2+1/2}{1+1/2}\frac{21}{19}\frac{m_O}{v_O}
 <1.184\frac{m_O}{v_O}.
\]
Here the loss ratio uses $B_N\ge B_S/2$. The ratio $m_O/C_-$ is at
most $\omega/.95$ by the $5\%$ signal-noise bound. Thus for every
$0<\gamma\le1.5\gamma_*$ the displacement is less than $.089$
and the sufficient-decrease ratio $tH(Y)/C(Y)$ is less than $1.88$.
The endpoint cap is also inactive by the bound in the linear proof
with $1.5$ replaced by $1.5(1.184)$. The interval centered at
$\gamma_*$ with relative half-width
$\tfrac12\sqrt{1-G_A\widetilde v/\widetilde m^2}$ is therefore
inside the safe range, and its minimum linear gap is
$3(\widetilde m^2/\widetilde v-G_A)/8$.
\end{proof}

The sufficient family shows how the dispersion contrast can favor
LoopAda even with observation noise on signal predictors and fewer
nuisance predictors than signal predictors. The controller uses the
entire loss. Its common multiplier changes all coordinates when a
single response changes, so the exact moment calculation treats the
coordinates jointly.

\subsection{Regression with Two Predictors and Local Robustness}

\begin{corollary}[Recovery of the first-update result for two predictors]\label{gen:cor:original}
Take $S=\{1\}$, $N=\{2\}$, $w^\star=(\mu,0)$, $\delta_1=0$,
$\sigma_2=\sigma>0$, $\kappa_1=1$, and $\kappa_2=\kappa>1$.
For $10^{-4}\le\mu\le.05$ and the original rate range, the linear case of
Corollary~\ref{cor:general-family} includes the original common $\gamma$
interval and its $3\Delta/4$ margin below best fixed-rate Adam.
\end{corollary}
\begin{proof}
Write $A=\mu/(\mu+\epsilon)$, $B=\sigma/(\sigma+\epsilon)$,
$\tau=(\sigma+\epsilon)/(\kappa\sigma+\epsilon)$,
$J=A^2+B^2$, $H=A^2+\tau^2B^2$, $C=\mu A+\sigma\tau B$,
and $F_0=(\mu^2+\sigma^2)/2$. Then
\[
 a_-=a_+=F_0/C,\quad \rho=1,\quad
 m_A=m_O=\mu A,\quad v_A=J,\quad v_O=H.
\]
Hence
\[
 \gamma_c=\frac{\mu AC}{HF_0},\quad
 \tau_\gamma=\frac12B\sqrt{\frac{1-\tau^2}{J}},\quad
 G_O-G_A=\frac{\mu^2A^2B^2(1-\tau^2)}{JH}=2\Delta>0.
\]
The interval and margin coincide with those in Supplementary Theorem~\ref{thm:occ-first-risk}. Removing the auxiliary state for the first predictor from \eqref{gen:eq:block} gives the block in \eqref{eq:occ-risk-block}.
\end{proof}

\begin{proposition}[Persistence under small signal leakage and parameter perturbations]\label{gen:prop:open}
Fix a finite-dimensional linear instance of Corollary~\ref{cor:general-family}
and its closed interval $I$. There is a nonzero neighborhood of its
numerical parameters, with $d,T,\epsilon$, the moment coefficients and
the complete update rule fixed, relative to the admissible parameter domain
(including $0<r\le1$, nonnegative signal-noise amplitudes, positive nuisance
amplitudes and $\kappa_j\ge1$), in which the same $I$ gives first-trial
acceptance and at least half its original risk margin. The perturbations
may include small nonzero means on the nuisance coordinates, with the
training and population means changed together. Either
\eqref{gen:eq:alignment} or \eqref{gen:eq:family} provides sufficient
conditions for a base model and a suitable interval $I$.
\end{proposition}
\begin{proof}
There are finitely many sign realizations. At the base instance every safeguard inequality proved above has strict slack uniformly over compact $I$; the norm and slope denominators are positive. The first-trial outputs and all guard quantities are consequently continuous in the numerical parameters and in $\gamma$ on a neighborhood of this compact set. Uniform continuity preserves their strict inequalities.

For any perturbed finite-support model, Adam's fixed-scalar oracle risk is
\[
 \frac12\|w^{\star\prime}\|^2-
 \frac{\{(w^{\star\prime})^\top\E p'(Y')\}_+^2}{2\E\|p'(Y')\|^2},
\]
which is continuous because its denominator is positive at the base point. Expected LoopAda risk is also a finite sum of continuous functions. The original gap is at least $M_I>0$ uniformly on $I$, so a sufficiently small common parameter neighborhood preserves a gap of at least $M_I/2$. Nuisance means can be varied while keeping their additive Rademacher amplitudes fixed; sufficiently small changes also preserve their nonzero response magnitudes. The argument establishes a positive common neighborhood with an implicit radius.
\end{proof}

\subsection{Illustrations on the Orthogonal Design}\label{app:nine-predictor}

For a sign vector $s\in\{-1,1\}^d$, let $Y(s)$ denote the corresponding
response vector in \eqref{gen:eq:Y}. Every expectation below is the finite
average $\E u(Y)=2^{-d}\sum_{s\in\{-1,1\}^d}u\{Y(s)\}$. Completing
the square in the expected risks gives
\[
 \begin{aligned}
 \inf_{\eta\ge0}\E\mathcal R\{w_A^+(\eta)\}
   &=\tfrac12\|w^\star\|^2-\tfrac12G_A,\\
 \E\mathcal R\{w_O^+(\gamma_*)\}
   &=\tfrac12\|w^\star\|^2-\frac{\widetilde m^2}{2\widetilde v},
 \qquad \gamma_* = \frac{\widetilde m}{\widetilde v}.
 \end{aligned}
\]
Under \eqref{gen:eq:safe}, the scale $\gamma_*$ lies in the accepted
first-trial range. When $\widetilde m^2>\widetilde vG_A$, the two roots
of the exact gap $M(\gamma)$ are
\begin{equation}\label{gen:eq:exact-scale-roots}
 \gamma_\pm=
 \frac{\widetilde m\pm\sqrt{\widetilde m^2-\widetilde vG_A}}
      {\widetilde v}.
\end{equation}
Consequently, every scale in
$(\gamma_-,\gamma_+)\cap(0,\Gamma_{\rm safe})$ gives an accepted first
trial with positive risk gap. The examples below specialize these finite
averages and quadratic roots.

\paragraph{Two signal predictors and one nuisance predictor.}

At the count boundary $2d_N=d_S$, take $T=8$, $r=8^{-1/2}$,
$w^\star=(.01,.01,0)$, signal-noise amplitudes $(.0005,.0005)$,
nuisance-noise amplitude $.01$, and dispersion factors $(1.2,1.2,6)$.
The finite averages defining $\widetilde m$ and $\widetilde v$ give
$\gamma_*\approx1.436$.
The expected excess risks are approximately $1.983\times10^{-6}$
for LoopAda and $3.333\times10^{-5}$ for best fixed-rate Adam.
These values are rounded evaluations of rational averages, and all eight
noise patterns satisfy the first-trial acceptance inequalities.

\paragraph{Nine predictors.}

Take $T=8$, $r=8^{-1/2}$, $d_S=3$, $d_N=6$, and
\[
 \begin{aligned}
  w^\star_S&=(.010,.008,.006),&\delta_S&=10^{-5}(30,32,12),\\
 \kappa_S&=(1.05,1.10,1.15),&
 \sigma_N&=(.020,.025,.030,.035,.040,.045),\\
 &&\kappa_N&=(6.5,7,8,9,10,11).
 \end{aligned}
\]
All other coefficients are zero. The nonzero coefficients have observation
noise, and the dispersion factors vary within both coefficient groups.
This model satisfies Corollary~\ref{cor:general-family}, with
\[
 \begin{gathered}
  \rho\approx.9663,\quad G_O\approx1.754\times10^{-4},\quad
  G_A\approx6.400\times10^{-5},\\
  \gamma_c\approx.1032,\qquad M_0\approx4.177\times10^{-5},\\
  \inf_{\eta\ge0}\E\mathcal R(w_A^+(\eta))\approx6.800\times10^{-5}.
 \end{gathered}
\]
The endpoints of $I_0$ are approximately $.06205$ and $.1443$.
The exact controller moments give a maximizer $\gamma_*\approx.1086$
and a maximum gap of approximately $6.191\times10^{-5}$.
The roots in \eqref{gen:eq:exact-scale-roots} satisfy
$\gamma_-\approx.02042$ and $\gamma_+\approx.1968$. Both roots lie below
$\Gamma_{\rm safe}$, so LoopAda has strictly lower risk for
$\gamma\in(\gamma_-,\gamma_+)$. At the roots the risks are equal.
The risks and margins at the endpoints and center of $I_0$, rounded to
four significant figures, are
\begin{center}
\begin{tabular}{lrr}
\toprule
$\gamma$ & Expected risk & Margin below oracle Adam\\
\midrule
 $\gamma_c(1-\tau_\gamma)$ & $2.334\times10^{-5}$ & $4.466\times10^{-5}$\\
 $\gamma_c$ & $6.322\times10^{-6}$ & $6.168\times10^{-5}$\\
 $\gamma_c(1+\tau_\gamma)$ & $1.622\times10^{-5}$ & $5.178\times10^{-5}$\\
\bottomrule
\end{tabular}
\end{center}
Nuisance sign changes preserve the squared risk and every controller and
safeguard quantity. The expectation therefore reduces to the eight
signal-noise patterns, each representing $2^6$ configurations of nuisance
signs. Throughout the expanded scale interval, the first trial is accepted
without clipping or fallback and has the form \eqref{gen:eq:exact-output},
including the endpoints and center of $I_0$ displayed above. Rational
bounds enclose the two roots and establish the safeguard inequalities
throughout this interval, as well as the inequalities used for
\eqref{gen:eq:family}.
\section{Proof of Theorem~\ref{thm:occ-window}}\label{app:occ-window}
Sections~\ref{app:window-invariants} and~\ref{app:window-acceptance}
establish the moment invariants and accepted update recursion in the
linear model. Sections~\ref{app:window-bounds} and~\ref{app:window-linear}
derive uniform risk bounds over the theorem's parameter and rate ranges
and extend the comparison to both noise signs.
Section~\ref{app:nonlinear-window} proves persistence under small smooth
perturbations and gives an explicit interval for the cubic model.
The linear and cubic bounds are obtained in real arithmetic using
outward interval enclosures with exact integer or rational endpoints.

\subsection{Moment Invariants in the Linear Model}
\label{app:window-invariants}
At $\lambda=0$, the occurrence coefficients in \eqref{eq:occ-risk-block} satisfy
\[
 \sum_t c_{t,j}=1,\qquad
 T\sum_t c_{t,1}^2=1,\qquad
 T\sum_t c_{t,2}^2=\kappa^2.
\]
The auxiliary state is independent of the trainable tensor, so $g_{t,j}=c_{t,j}g_j$ holds at every parameter point. Consequently the pooled and aggregate second moments satisfy the exact invariant
\begin{equation}\label{eq:occ-window-invariant}
 q_{k,1}=v_{k,1},\qquad q_{k,2}=\kappa^2v_{k,2}
\end{equation}
by induction from zero, since they use the same second-moment coefficient. The invariant relates the two second moments computed along a single LoopAda trajectory.

For either method, with its own parameter and moment history, put $y_\chi=(\mu,\chi\sigma)$ and use
\begin{align*}
 g_k&=w_{k-1}-y_\chi,\\
 m_k&=\tfrac9{10}m_{k-1}+\tfrac1{10}g_k,\\
 v_k&=\tfrac{999}{1000}v_{k-1}+\tfrac1{1000}g_k^{\odot2},\\
 \widehat m_k&=m_k/\{1-(9/10)^k\},\\
 \widehat v_k&=v_k/\{1-(999/1000)^k\}.
\end{align*}
Adam uses $a_k=\widehat m_k/(\sqrt{\widehat v_k}+\epsilon)$ and $w_k^A=w_{k-1}^A-\eta a_k$. All operations in these definitions are coordinatewise when appropriate. For LoopAda, \eqref{eq:occ-window-invariant} gives
\begin{equation}\label{eq:occ-window-u}
 u_{k,1}=\frac{\widehat m_{k,1}}{\sqrt{\widehat v_{k,1}}+\epsilon},\qquad
 u_{k,2}=\frac{\widehat m_{k,2}}{\kappa\sqrt{\widehat v_{k,2}}+\epsilon}.
\end{equation}
These recursions retain the finite denominator constant $\epsilon=10^{-8}$.

\subsection{First-Trial Acceptance and the Reduced Recursion}
\label{app:window-acceptance}
Suppress the update index and define
\[
 U=\|u\|^2,\quad A=\|a\|^2,\quad
 \rho=\sqrt{A/U},\quad G=g^\top u,\quad
 F=\|g\|^2/2,\quad h=\gamma F/G.
\]
The interval bounds below establish, at every update $1,\ldots,16$,
\begin{align}
 U&>10^{-60},\qquad A>10^{-40},\nonumber\\
 s&=r\rho G>10^{-30},\qquad \|w\|^2<1,\nonumber\\
 h^2U&<.01,\qquad \alpha_{\rm loss}=\gamma F/s<10^{19},\nonumber\\
 \tfrac{19}{20}\gamma F-\tfrac12h^2U&>0.\label{eq:occ-window-guards}
\end{align}
The first two conditions give active norm matching, $d=\rho u$, and $\|d\|=\sqrt A>10^{-20}$. The slope condition excludes fallback and the slope floor. Before clipping, the physical movement is
$\alpha_{\rm loss}rd=hu$; hence $\|w\|^2<1$ and $h^2U<.01$ imply that the block movement cap is inactive. The empty endpoint group has cap $10^{19}$, which is also inactive. Finally,
\[
 F(w)-F(w-hu)=hG-\tfrac12h^2U
 =\gamma F-\tfrac12h^2U
 >\tfrac1{20}\alpha_{\rm loss}s.
\]
The first trial is therefore accepted, and the optimizer has the exact update
\begin{equation}\label{eq:occ-window-reduced}
 w_k=w_{k-1}-\frac{\gamma F(w_{k-1})}{g_k^\top u_k}u_k
\end{equation}
through step 16. The strict inequalities in \eqref{eq:occ-window-guards} establish first-trial acceptance, so the attempt and update indices coincide.

Uniformly over the complete parameter box, every step satisfies $s>.006$, $U>.0105$, $A>.994$, $h^2U<1.964\times10^{-6}$, $\alpha_{\rm loss}<.00318$, and $\tfrac{19}{20}\gamma F-\tfrac12h^2U>1.72\times10^{-5}$. The displayed decimals are conservative roundings of exact rational bounds.

\subsection{Uniform Risk Bounds over the Adam Rate Range}
\label{app:window-bounds}
Represent an interval as $[L/S,U/S]$ with integer endpoints and $S=2^{128}$. Rational inputs are rounded outward to this grid. Addition and subtraction combine endpoints exactly. Multiplication takes the extrema of the four endpoint products and divides by $S$, rounding down for the lower endpoint and up for the upper endpoint. Division by a strictly positive interval similarly takes the extrema of the four ratios $aS/b$ and rounds outward. Squaring uses lower endpoint zero when the input straddles zero and otherwise the smaller endpoint square; its upper endpoint is the larger endpoint square. For a nonnegative interval, square root has integer endpoints
\[
 \left\lfloor\sqrt{LS}\right\rfloor,
 \qquad
 \left\lceil\sqrt{US}\right\rceil,
\]
computed by integer square root and an exact perfect-square test.

Each operation encloses its exact real value by monotonicity and endpoint extrema. Induction therefore encloses every moment, parameter, loss, and safeguard expression. Repeated parameter dependence can widen the intervals while preserving inclusion. Every denominator has a strictly positive lower bound. Propagating the whole parameter box \eqref{eq:occ-window-box} through each recursion includes simultaneous variation of all four parameters.

For Adam, start with the dyadic enclosure of $[0,1/10]$ and propagate the full recursion for each rate interval through step 16. Include an interval in the final covering when its risk lower bound is at least $3/125000$ at each $K=6,\ldots,16$; otherwise bisect it at its integer midpoint. The resulting covering consists of 803 closed intervals, with maximum bisection depth 13. Exact rational comparison places every risk lower endpoint over this covering and all eleven updates above $3/125000$.

For LoopAda, propagating \eqref{eq:occ-window-u}--\eqref{eq:occ-window-reduced} under \eqref{eq:occ-window-guards} and comparing exact rational endpoints bounds the risk strictly below $3/200000$ over $K=6,\ldots,16$. These rational thresholds give the exact risk comparison used in the theorem.

A second enclosure uses exact rational endpoints, outward rounding to $2^{-160}$, exact rational bias-correction powers, and integer square roots. The rate intervals cover $[0,1/10]$ without gaps, and all eleven risk bounds hold on every interval. Propagating the original pooled-moment recursion gives enclosures for the normalized direction, slope, caps, trial, and training-loss decrease. The trial enclosure and \eqref{eq:occ-window-reduced} contain the same exact update under the safeguard inequalities, so their intersection is also a valid enclosure. Both constructions apply to the complete parameter box.

\subsection{Noise-Sign Symmetry and Linear Risk Separation}
\label{app:window-linear}
Changing $\chi$ negates coordinate 2 of $y_\chi,w,g,m,a,u,d$, leaves
coordinate 1 unchanged, and leaves both second moments unchanged.
Induction preserves this relation for either optimizer. Norms, slopes,
losses, caps, acceptance decisions, and $\mathcal R$ are invariant.
The positive-sign certificate therefore applies to both signs. The
complete parameter and rate covering gives a LoopAda risk of at most
$1.5\times10^{-5}$ and an Adam risk of at least $2.4\times10^{-5}$,
uniformly and separately over all iterates from 6 through 16.
Thus the risk ratio is at most $1.5/2.4=5/8$, and the absolute gap
is at least $9\times10^{-6}$.

\subsection{Nonlinear Perturbations and the Cubic Case}
\label{app:nonlinear-window}
On the same design, use $\psi_{\lambda,j}$ from \eqref{gen:eq:response-link}
and $Y_\lambda=(\psi_{\lambda,1}(\mu),\chi\sigma)$.
At a current iterate $w$ put $e_j=\psi_{\lambda,j}(w_j)-Y_{\lambda,j}$
and
\[
 b_{t,j}(w_j)
 =c_{t,j}\{1+\lambda\varphi'(\sqrt 2\,c_{t,j}w_j)\}.
\]
The gradient and the input to the pooled second moment in coordinate
$j$ are, respectively,
\begin{equation}\label{eq:window-nonlinear-gradients}
 g_j=e_j\sum_t b_{t,j}(w_j),\qquad
 E_j=T e_j^2\sum_t b_{t,j}(w_j)^2.
\end{equation}
Use $F_\lambda(w)=\sum_j e_j^2/2$ in the step-size rule and trial
loss. The moment recursions in Algorithm~\ref{alg:occurrence-sgd}
use $g_j$ and $E_j$ evaluated along each optimizer's own trajectory,
with $E_j$ used only by LoopAda.

\paragraph{Persistence of the risk gap.}
The gradients and moment recursions are continuous in the perturbation,
the model parameters, and the current state. Square roots are
continuous on nonnegative inputs and the denominator constant
$\epsilon$ is positive. Hence Adam's iterates are continuous for
each finite update count. For LoopAda, the strict norm, descent,
cap, and first-trial acceptance conditions hold at zero perturbation
for every parameter tuple and noise sign. At step one
these conditions persist by continuity of the trial and loss.
Inductively, continuity of the accepted iterate and updated moments
makes the same argument applicable at the next step. Compactness
of $\mathcal P$ and finiteness of the steps and signs yield a common
positive radius for all sixteen updates. The risk difference between
Adam and LoopAda is then continuous on
$[0,\bar\lambda]\times\mathcal P\times[0,.1]$ for each step and sign,
where $\bar\lambda$ can be chosen within this radius.
At zero perturbation the separate upper and lower risk bounds apply at
every iterate in the range. Uniform continuity on the same compact
sets lets us reduce $\bar\lambda$ so that each risk changes by at most
$3\times10^{-7}$. LoopAda's risk is then at most $1.53\times10^{-5}$
and Adam's at least $2.37\times10^{-5}$, at any pair of iterates.
The risk ratio is therefore at most $1.53/2.37=51/79<.65$, and the
absolute gap at least $8.4\times10^{-6}>7.9\times10^{-6}$.

\paragraph{An explicit interval for the cubic block.}
For $\varphi(u)=u^3$, put $S_{m,j}=\sum_t c_{t,j}^m$.
The normalized response becomes
$\psi_{\lambda,j}(w_j)=w_j+2\lambda S_{3,j}w_j^3$, and
\eqref{eq:window-nonlinear-gradients} gives
\[
 E_j=T e_j^2\bigl(S_{2,j}+12\lambda w_j^2 S_{4,j}
                  +36\lambda^2 w_j^4 S_{6,j}\bigr).
\]
The equal coefficients for predictor 1 give $q_{k,1}=v_{k,1}$.
For predictor 2 we update $q_{k,2}$ from $E_2$ at every step.
The occurrence coefficients are symmetric about $1/T$, so only
even powers of their centered offsets contribute when evaluating
$S_{m,j}$ by the binomial formula. This allows interval evaluation
using the squared spacing $b_\kappa^2$ in \eqref{eq:occ-risk-block}.

The same outward rational enclosures extend over the full interval
$[0,10^{-3}]$ of perturbation strengths and the full parameter box
at each update. They establish the norm and slope thresholds, both
caps, and sufficient decrease for the nonlinear trial loss.
For Adam, subdividing the linear rate intervals until the required
risk lower bounds hold gives
1,240 intervals that cover the entire Adam rate range. Over updates
6 through 16, the exact rational upper endpoints for LoopAda are at most
$1.4362\times10^{-5}$ and the lower endpoints for Adam at least
$2.2276\times10^{-5}$. For this cubic model, the predictors are odd
functions of their respective weights and their derivatives are even.
Thus the sign symmetry of the gradients, moments, acceptance tests,
and risks again covers both noise signs. The risk ratio is at most
$1.4362/2.2276<.65$, and the absolute gap at least
$7.914\times10^{-6}>7.9\times10^{-6}$, for any pair of iterates
throughout the cubic interval.

\par\phantomsection\label{proof:occ-window-end}
This completes the proof of Theorem~\ref{thm:occ-window}.\hfill\qedsymbol
\section{Experimental Methods and Results}\label{sec:experimental-methods}
\subsection{Simulation Study in Linear Regression}\label{sec:signal-placement-supp}
The fixed block is the realization in \eqref{gen:eq:block} with $d=T=8$,
$\kappa_j=1.1$ on the first four coordinates and $\kappa_j=8$ on the last four.
There is one trainable tensor and the input and output maps are fixed. All
targets have four coefficients equal to $.01$, with the rest zero. We vary the number of nonzero coefficients in the high-dispersion group
from zero to four, keeping the norm of the true coefficient vector fixed.

In the orthogonal design of Section~\ref{app:general-risk}, the additive
training-noise amplitude is $.0003$ on signal coordinates and $.02$ on
coordinates with zero coefficients. We enumerate all $2^8$ independent training-sign
vectors for each placement. Each LoopAda first update accepts its first
trial and agrees numerically with the first-step formula.
The common controller scale $.45$ lies in the theorem's guaranteed interval
when all four signal coordinates are in the low-disagreement group. The
Adam rate is its population optimum for a fixed scalar first step, computed
from the exact noise average. We also compute the optimal LoopAda scale
$\widetilde m/\widetilde v$ at each placement; its first
trial is accepted without clipping or fallback for all 256 noise
patterns. From zero through four signal predictors in the
high-dispersion group, the optimized LoopAda-to-Adam risk ratios are
$.037$, $.792$, $1.365$, $1.755$, and $1.963$.
The risks and first-trial conditions follow from exact rational arithmetic.

In the correlated design, each problem has 64 rows drawn from a centered
Gaussian with unit marginal variances and pairwise correlation $.25$. The
responses are linear with independent centered Gaussian noise of standard
deviation $.06$. Permuting the columns and the true coefficients together
preserves the responses. For each placement, we estimate the best fixed
scalar Adam rate from the population-risk quadratic averaged over 4,096
independent calibration problems. A second sample of 4,096 training problems
evaluates both optimizers using the known population covariance for risk.
LoopAda uses the same scale $.45$.
The paired Monte Carlo standard errors of the risk difference at the two
endpoint placements are $1.04\times10^{-6}$ and $2.23\times10^{-6}$, respectively.

\subsection{Nonlinear Regression with Repeated Measurements}
\label{sec:nonlinear-regression-supp}
For each observation and predictor $j$, draw $x_{1,j}\sim N(0,1)$
and let $x_{t,j}=\rho_j x_{t-1,j}+\sqrt{1-\rho_j^2}\,z_{t,j}$ for
$t=2,\ldots,8$, with independent $N(0,1)$ innovations $z_{t,j}$. The
predictor trajectories are independent across $j$. The lag-one
correlations are $.85$ for the first four predictors and $-.85$ for the
other 60. The conditional mean is the output of the recurrence in
Section~\ref{sec:nonlinear-regression} with input weights
$(.55,-.9,1.2,-1.5)$ on the first four predictors, zero input weights
on the others, all feedback weights equal to $.3$, and zero intercept.
The training and validation responses add independent $N(0,1)$ noise.

In each fit the initial intercept is zero. The initial input and feedback
weights are independent Gaussians with standard deviation $.03$ and means
zero and $.3$, respectively. We minimize half the minibatch MSE, with
minibatches of 64 observations sampled with replacement from 128 training
observations. Both optimizers make 120 update attempts and use a
shared-block rate multiplier of one.
Model selection minimizes MSE on 256 independent validation observations,
evaluating the initial fit and every sixth attempt.
Ties select the earlier iterate.

For development, a separate set of 1,024 validation observations per
replication scores the validation-selected fit. We choose rates by mean
score over 32 development replications. The grids use multipliers $2^j$
for $j=-2,\ldots,3$, with base values $.01$ for Adam and $1/64$ for
the LoopAda scale. The selected values are $.04$ for Adam and $1/32$
for LoopAda, both interior to their grids. The 128 paired evaluation
replications use randomizations independent of development and the pilot.
Each test set consists of 4,096 independent predictor draws with known
conditional means. The interval for the ratio of mean excess prediction
MSEs resamples whole paired replications 12,000 times, conditional on the
selected rates. Both optimizers accepted all 120 attempts in every
evaluation replication.

At each probed attempt, let $g_{r,t,j}$ denote the loss-gradient
contribution of block application $t$ for input weight $j$ on the current
minibatch of replication $r$, evaluated before the update. For each
predictor group $C$, the disagreement-energy fraction is
\[
 1-\frac{\sum_r\sum_{j\in C}(\sum_{t=1}^{T}g_{r,t,j})^2}
          {T\sum_r\sum_{j\in C}\sum_{t=1}^{T}g_{r,t,j}^2},
 \qquad T=8.
\]
This fraction measures the disagreement energy as a share of total
per-application gradient energy. Its endpoints are zero for identical
contributions and one for complete cancellation, provided the denominator
is positive. The four probes precede attempts $1,25,61,115$. On Adam's
training trajectories, the disagreement fractions range from $.984$ to
$.988$ for nuisance predictors and from $.081$ to $.563$ for signal
predictors. LoopAda's corresponding ranges are $.981$--$.985$ for nuisance
predictors and $.081$--$.531$ for signal predictors.

\paragraph{Pooling comparison.}
The pooling control repeats the fit with gradient pooling disabled, as
defined in Section~\ref{app:controlled-ablation}. It uses the same grid
and development replications for tuning. Its selected scale is $1/32$
and its mean excess MSE is $.214$. The ratio of LoopAda's mean error to
the control's is $.344$, with a paired bootstrap interval of
$[.324,.365]$ adjusted by Bonferroni for simultaneous 95\% confidence
across the two LoopAda comparisons. The corresponding adjusted interval
for LoopAda relative to Adam is $[.337,.382]$.

\subsubsection{Comparison with Adam Variants}
\label{sec:nonlinear-strong-baselines-supp}
AdamW and Schedule-Free AdamW use the same 32 development replications
and validation criterion as the preceding comparison. Each variant searches
seven learning rates, $.0025\,2^j$ for $j=0,\ldots,6$, and five values
of block weight decay, $(0,.01,.1,.5,2)$. The intercept is exempt from
decay. For AdamW we compare a constant rate with a 12-attempt warmup
followed by cosine decay to one tenth of the base rate. For Schedule-Free
AdamW we compare zero and 12 warmup attempts. Both use $\beta$ settings
$(.9,.999)$ and denominator constant $10^{-8}$. Schedule-Free AdamW
uses averaged evaluation weights for model selection and prediction.

AdamW selects a constant learning rate of $.04$ and weight decay of $.1$.
Schedule-Free AdamW selects a rate of $.01$, zero weight decay, and
no warmup. Adam and LoopAda keep the settings from the original development
study. The 128 evaluation replications are disjoint from the preceding
128 and from development. Within each replication all methods use the
same data, initialization, and minibatches. The intervals in
Table~\ref{tab:nonlinear-original-baselines} use 12,000 paired bootstrap
resamples with a Bonferroni adjustment across the three ratios.
\begin{table}[htbp]
\centering\small
\caption{Nonlinear prediction with Adam variants}
\label{tab:nonlinear-original-baselines}
\begin{tabular}{@{}lrc@{}}
\toprule
Optimizer & Excess MSE & Ratio [interval]\\
\midrule
LoopAda & .0710 & ---\\
Adam & .1979 & $.359\,[.335,.383]$\\
AdamW & .1965 & $.361\,[.337,.387]$\\
Schedule-Free AdamW & .2221 & $.320\,[.298,.342]$\\
\bottomrule
\end{tabular}
\par\smallskip
\begin{minipage}{\linewidth}\footnotesize
Ratios are LoopAda/comparator mean excess MSE. Brackets give paired
bootstrap intervals with nominal 95\% simultaneous coverage across three
comparisons.
\end{minipage}
\end{table}

\FloatBarrier
\subsubsection{Training Horizons and Temporal Dependence}
\label{sec:nonlinear-followup-supp}
We evaluate repeated-measurement prediction across training budgets
and temporal correlations of nuisance predictors. Write $(\rho_S,\rho_N)$
for the signal and nuisance AR(1) correlations. We consider
$F=(.85,-.85)$ at maximum budgets $H=120,480,1920$ and
$M=(.85,-.50)$, $E=(.85,.85)$, and $R=(-.85,.85)$ at $H=480$.
The conditional-mean function, shared network, noise variance,
initialization law, and loss are as specified above. The first three
correlation settings keep the signal process fixed. Setting $R$ changes
both the signal process and its variance relative to observation noise.

\paragraph{Data and selection.}
Each fit uses 128 training observations, 256 independent observations for
model selection, and minibatches of size 64 sampled with replacement.
Model selection compares the initial model and every sixth attempted
update through $H$, choosing the earliest minimizer of noisy validation
MSE. For each method and setting, a further 1,024 independent noisy
validation observations score the selected fit in each of 16
development replications. Configurations minimize the average of these
scores, ranking configurations with fewer failed development fits first.
The selected configurations are evaluated on 64 independent replications,
each with 4,096 new test inputs and their true conditional means.
Development and evaluation use separate randomizations, also distinct
from those in the preceding studies.

Within each replication, we use the same data and initialization across
methods and training budgets. All methods see the same minibatches in the
same order up to their respective budgets. Across correlation settings,
we reuse the same Gaussian innovations and observation-noise draws. These common replications induce dependence
among horizon and setting comparisons. Noisy responses determine model
and configuration selection; true conditional means determine test risk.

\paragraph{Optimizer selection.}
The pooling control disables pooling in LoopAda and retains the same
norm matching, movement caps, slope fallback, and step-size rule. Both
use a shared-block multiplier of one. Their initial scales are $2^j/64$ for
$j=-2,\ldots,3$; Adam initially uses $.01\,2^j$ over the same exponents.
AdamW and Schedule-Free AdamW each initially search 70 configurations:
seven rates $.0025\,2^j$, $j=0,\ldots,6$, five block-weight decays
$(0,.01,.1,.5,2)$, and two schedules. The intercept is exempt from decay.
AdamW compares constant rates with a warmup over $H/10$ attempts followed
by cosine decay to one tenth of the base rate. Schedule-Free AdamW
compares zero warmup with $H/10$ warmup attempts and uses averaged
weights for validation and prediction. Moment coefficients are
$(.9,.999)$, with denominator constant $10^{-8}$.

Up to two additional development rounds extend a selected boundary rate
by a factor of two, within $[1/6400,.64]$, and double a selected
largest positive decay up to 8. The two LoopAda variants share any rate
extension. Each horizon is tuned separately. Table~\ref{tab:nonlinear-followup-settings}
gives the selected configurations, including remaining boundary choices.
LoopAda and the pooling control select interior rates in every setting.
Selected schedules are constant except for AdamW in $R$, which uses
warmup and cosine decay.

\begin{table}[htbp]
\centering\small
\caption{Selected optimizer configurations for nonlinear regression}
\label{tab:nonlinear-followup-settings}
\setlength{\tabcolsep}{4pt}
\begin{tabular}{@{}crccccc@{}}
\toprule
Setting & $H$ & LoopAda & \shortstack{Pooling\\control} & Adam & AdamW & Schedule-Free\\
\midrule
F & 120  & $1/4$ & $1/32$ & $.02$ & $(.04,.1)$ & $(.02,0)$\\
F & 480  & $1/4$ & $1/32$ & $.02$ & $(.04,.1)$ & $(.005,.01)$\\
F & 1920 & $1/4$ & $1/32$ & $.02$ & $(.04,.1)$ & $(1/800,0)$\\
M & 480 & $1/8$ & $1/32$ & $.01$ & $(.64,2)^{\dagger}$ & $(.005,.5)$\\
E & 480 & $1/256$ & $1/128$ & $1/800$ & $(.08,2)$ & $(.64,2)^{\dagger}$\\
R & 480 & $1/512$ & $1/512$ & $(1/1600)^{\dagger}$ & $(1/1600,8)^{\ddagger}$ & $(1/1600,8)^{\ddagger}$\\
\bottomrule
\end{tabular}
\par\smallskip
\begin{minipage}{\linewidth}\footnotesize
LoopAda columns: controller scales; Adam: learning rate; AdamW/Schedule-Free:
(rate, decay). $\dagger$: boundary rate; $\ddagger$: boundary rate and
largest searched decay.
\end{minipage}
\end{table}

\paragraph{Prediction risk.}
Table~\ref{tab:nonlinear-followup-means} gives all mean excess MSEs.
For each setting and horizon, we resample the 64 paired replications 20,000 times and
calculate the ratio of arithmetic mean errors in every resample.
The intervals in Table~\ref{tab:nonlinear-followup-ratios} use quantiles
$.05/(2\cdot24)$ and $1-.05/(2\cdot24)$, giving a nominal simultaneous
95\% confidence level across the six combinations of setting and horizon
and four comparators.
They describe variation over new simulated samples and training randomization,
conditional on the selected configurations. The Bonferroni adjustment
accommodates dependence across horizons and correlation settings.

\begin{table}[htbp]
\centering\small
\caption{Mean excess MSE across correlation structures and training horizons}
\label{tab:nonlinear-followup-means}
\setlength{\tabcolsep}{5pt}
\begin{tabular}{@{}crrrrrr@{}}
\toprule
Setting & $H$ & LoopAda & \shortstack{Pooling\\control} & Adam & AdamW & Schedule-Free\\
\midrule
F & 120 & $.0712\,(.0038)$ & $.2064\,(.0038)$ & $.1963\,(.0035)$ & $.1931\,(.0037)$ & $.2139\,(.0036)$\\
F & 480 & $.0712\,(.0038)$ & $.2064\,(.0038)$ & $.1963\,(.0035)$ & $.1931\,(.0037)$ & $.2138\,(.0036)$\\
F & 1920 & $.0712\,(.0038)$ & $.2064\,(.0038)$ & $.1963\,(.0035)$ & $.1931\,(.0037)$ & $.2120\,(.0036)$\\
M & 480 & $.1212\,(.0043)$ & $.3362\,(.0048)$ & $.3337\,(.0048)$ & $.3032\,(.0035)$ & $.3366\,(.0047)$\\
E & 480 & $.7708\,(.0109)$ & $.8173\,(.0127)$ & $.8221\,(.0117)$ & $.6370\,(.0113)$ & $.6049\,(.0108)$\\
R & 480 & $.0795\,(.0011)$ & $.0790\,(.0011)$ & $.0790\,(.0011)$ & $.0638\,(.0009)$ & $.0636\,(.0008)$\\
\bottomrule
\end{tabular}
\par\smallskip
\begin{minipage}{\linewidth}\footnotesize
$F,M,E,R$ denote $(\rho_S,\rho_N)=(.85,-.85),(.85,-.50),(.85,.85),(-.85,.85)$,
respectively. The pooling control uses the squared summed gradient in its second
moment and is tuned separately. Parentheses give Monte Carlo standard
errors across 64 evaluation replications, conditional on the selected
configurations.
\end{minipage}
\end{table}

\begin{table}[htbp]
\centering\small
\caption{Relative prediction risk across correlations and training horizons}
\label{tab:nonlinear-followup-ratios}
\setlength{\tabcolsep}{4pt}
\begin{tabular}{@{}crcccc@{}}
\toprule
Setting & $H$ & \shortstack{Pooling\\control} & Adam & AdamW & Schedule-Free\\
\midrule
F & 120 & $.345\,[.297,.404]$ & $.363\,[.313,.425]$ & $.369\,[.320,.427]$ & $.333\,[.287,.390]$\\
F & 480 & $.345\,[.295,.403]$ & $.363\,[.310,.424]$ & $.369\,[.317,.427]$ & $.333\,[.286,.390]$\\
F & 1920 & $.345\,[.297,.405]$ & $.363\,[.311,.427]$ & $.369\,[.319,.429]$ & $.336\,[.290,.394]$\\
M & 480 & $.360\,[.329,.395]$ & $.363\,[.330,.400]$ & $.400\,[.361,.443]^{\dagger}$ & $.360\,[.328,.396]$\\
E & 480 & $.943\,[.926,.962]$ & $.938\,[.927,.947]$ & $1.210\,[1.170,1.250]$ & $1.274\,[1.231,1.317]^{\dagger}$\\
R & 480 & $1.006\,[.999,1.013]$ & $1.006\,[.998,1.014]^{\dagger}$ & $1.246\,[1.201,1.290]^{\ddagger}$ & $1.251\,[1.211,1.290]^{\ddagger}$\\
\bottomrule
\end{tabular}
\par\smallskip
\begin{minipage}{\linewidth}\footnotesize
Ratios are LoopAda/comparator mean excess MSE; paired bootstrap intervals
have nominal 95\% simultaneous coverage across 24 comparisons. Boundary
symbols refer to comparator settings in
Table~\ref{tab:nonlinear-followup-settings}.
\end{minipage}
\end{table}

In $F$ and $M$, LoopAda has lower excess MSE than each comparator in all
64 pairs. In $E$, its error is $5.7\%$ below that of the pooling control,
but $21.0\%$ and $27.4\%$ above AdamW's and Schedule-Free AdamW's.
All three adjusted intervals exclude one. In $R$, the adjusted pooling
interval includes one, while AdamW and Schedule-Free AdamW give lower
error. AdamW in $M$ and Schedule-Free AdamW in $E$ select upper rate
boundaries; the three Adam methods in $R$ select lower rate boundaries.
Both regularized methods in $R$ also select the largest searched decay.

\paragraph{Training duration and model selection.}
In $F$, LoopAda, its pooling control, Adam, and AdamW select the same
early iterates across budgets, giving identical errors for each method.
In $E$, several fits reach the last checkpoint
(Table~\ref{tab:nonlinear-followup-checkpoints}).

\begin{table}[htbp]
\centering\small
\caption{Validation-selected training duration in nonlinear regression}
\label{tab:nonlinear-followup-checkpoints}
\setlength{\tabcolsep}{5pt}
\begin{tabular}{@{}crccccc@{}}
\toprule
Setting & $H$ & LoopAda & \shortstack{Pooling\\control} & Adam & AdamW & Schedule-Free\\
\midrule
F & 120 & 12 [0] & 39 [0] & 36 [0] & 18 [0] & 72 [0]\\
F & 480 & 12 [0] & 39 [0] & 36 [0] & 18 [0] & 285 [0]\\
F & 1920 & 12 [0] & 39 [0] & 36 [0] & 18 [0] & 1161 [0]\\
M & 480 & 12 [0] & 42 [0] & 63 [0] & 294 [0] & 279 [0]\\
E & 480 & 354 [5] & 168 [0] & 285 [9] & 300 [1] & 177 [1]\\
R & 480 & 48 [0] & 51 [0] & 30 [0] & 87 [0] & 111 [0]\\
\bottomrule
\end{tabular}
\par\smallskip
\begin{minipage}{\linewidth}\footnotesize
Entries give median selected attempts; brackets count selections at the
maximum budget $H$ among 64 replications.
\end{minipage}
\end{table}

\paragraph{Signal variance and disagreement.}
The average Monte Carlo variance of the true conditional mean is
$1.136$ in $F$, $M$, and $E$, and $.0340$ in $R$.
Consequently, the reversed setting changes signal-to-noise ratio as well
as the ordering of temporal correlations.

For LoopAda and Adam, the disagreement-energy statistic above is measured
before attempts $1$, $\lfloor .2H\rfloor+1$,
$\lfloor .5H\rfloor+1$, and $\lfloor .95H\rfloor+1$ on the current
training minibatch. Initial signal/nuisance fractions are approximately
$(.076,.984)$ in $F$, $(.076,.951)$ in $M$, $(.077,.312)$ in $E$,
and $(.977,.309)$ in $R$. The two methods share these initial values.
The equal input autocorrelations in $E$ coexist with different gradient
disagreement for signal and nuisance predictors. Training and model
selection use the loss and validation criteria specified above; the
probes summarize the resulting trajectories.

\paragraph{Computation.}
Fits use one CPU thread. Measured fit CPU time includes controller trials,
validation, and restoration of the selected parameters. Data generation, test scoring,
and measured diagnostic overhead are excluded. At $H=1920$ in $F$,
mean fit CPU times are $7.965$, $7.331$, $4.936$, $5.169$, and $4.936$
seconds for LoopAda, the pooling control, Adam, AdamW, and Schedule-Free AdamW.
These are comparisons at matched attempted updates.
\FloatBarrier
\subsection{Four Real-Data Regression Case Studies}
\label{sec:real-regression-supp}

\subsubsection{Sources, variables, and grouping}

Table~\ref{tab:real-regression-data} distinguishes released source size
from evaluated sample size. Group definitions were fixed before fitting
and apply to both inner and outer partitions.
Stopping, configuration scoring, and test MSE all use equal observation
weights. Group separation controls the specified overlap but does not
establish statistical independence.

\begin{table}[htbp]
\centering\small
\caption{Data used in the four cases}
\label{tab:real-regression-data}
\setlength{\tabcolsep}{4pt}
\begin{tabularx}{\linewidth}{@{}lrrrrX@{}}
\toprule
Case & Source rows & Used rows & Inputs & Groups & Group key\\
\midrule
Student Portuguese & 649 & 649 & 58 & 637 & Profile tuple\\
CT Slices & 53,500 & 5,000 & 384 & 97 & Released patient code\\
Communities Crime & 1,994 & 1,993 & 100 & 46 & State\\
Diabetes & 442 & 442 & 10 & 442 & Exact input tuple\\
\bottomrule
\end{tabularx}
\par\smallskip
\begin{minipage}{\linewidth}\footnotesize
The input count gives the number of numerical design columns. Student Portuguese has 32 original
variables. CT uses one fixed sample selected without examining response
values. One community record lacks a required predictor.
\end{minipage}
\end{table}

\paragraph{Student Portuguese.}
The Portuguese-subject file of Student Performance~\cite{uci_student_performance}
contains 649 records from two schools. The response is the numerical final
grade \(G3\). Inputs are the 30 released background/school variables and the
earlier-period grades \(G1,G2\); fixed categorical vocabularies yield 58
numerical columns. All rows are retained.
The 13-field profile key follows the official matching script. Distinct
students can share a key. The task predicts final grades conditional on
earlier grades within the two source schools.

\paragraph{CT Slices.}
The axial-location data~\cite{uci_ct_slices} provide 53,500 slices,
384 bone/air histogram descriptors, and a relative axial-coordinate response.
The response interpolates anatomical landmarks annotated by hand. We use
the patient code for grouping and retain the \(-0.25\) marker for
histogram bins outside the image. Before fitting, we sampled 5,000 eligible
slices uniformly without replacement, independently of responses and model
scores. The sample covers all 97 patient codes in the released file.
The repository description reports 74 patients. We evaluate prediction
on the fixed sample by holding out groups defined by the released codes.

\paragraph{Communities Crime.}
Communities and Crime~\cite{uci_communities_crime} combines community-level
census, law-enforcement, and crime records. The response is the released
normalized violent-crimes-per-population variable.
A fixed exclusion list removes five identifier/split fields and 22 police
variables with extensive missingness, leaving 100 predictors.
Removing the one record with missing \texttt{OtherPerCap} leaves 1,993 rows.
The 46 state codes define groups and are excluded from inputs.
The released numeric variables had already been normalized and clipped
before we formed the state-level splits.

\paragraph{Diabetes.}
The original unscaled diabetes text data~\cite{diabetes_original_raw_mirror}
contain 442 records and a quantitative progression response measured one
year after baseline. Inputs are age, sex, body-mass index, blood pressure,
and six serum variables, retaining original textual precision.
All ten-dimensional input tuples are distinct and form 442 groups;
stable patient identifiers are not supplied.
We use these raw measurements without software-package global
standardization or polynomial expansion.

\paragraph{Preprocessing.}
Apart from the declared exclusions, fixed categorical coding, and CT
sampling, source values are retained without synthetic predictors or
new target transformations.
Each fit estimates predictor and response means and population standard
deviations from its fitting partition alone; a standard deviation at most
\(10^{-12}\) is replaced by one.
Predictions and squared errors are returned to the released response units.

\subsubsection{Network and optimizer settings}
\label{app:controlled-ablation}

The width-32 predictor is
\[
 h_0=\tanh(xE+a),\quad
 h_t=\tanh(h_{t-1}W+c),\quad t=1,\ldots,8,\quad
 f_w(x)=h_8v+o .
\]
The same \(W,c\) are used at every block application, and all parameters
are trainable. Encoder/readout entries have centered Gaussian initializations
with variances \(1/p\) and \(1/32\); \(W\) starts orthogonally and biases
start at zero.

LoopAda and Adam are the primary comparison. An auxiliary experiment
repeats the LoopAda fit with pooling disabled and tunes this setting
independently. The control replaces the depth-scaled sum of squared occurrence
contributions by the squared aggregate gradient in the shared-parameter
second moment, retaining LoopAda's first moment, tensorwise direction-norm
matching, loss-based controller, and acceptance rule.
This control assesses the contribution of pooling under the same step-size rule.
All methods use moment coefficients \((0.9,0.999)\), denominator offset
\(10^{-8}\), and no weight decay.
Each independently selects the shared-block factor \(8^{-\beta}\) with
\(\beta\in\{0,\tfrac12,1\}\).

Each fit minimizes half the minibatch MSE for 1,200 update attempts with
batches of 64 sampled uniformly with replacement.
Paired methods and candidates share splits, initial parameters, and
minibatch indices. We check the initial model, attempts \(1,5,10\), and
every multiple of 20 through 1,200, selecting the earliest checkpoint
with minimum stopping MSE.

\subsubsection{Nested selection and evaluation}

Five outer folds hold out groups. Within each outer training set,
four paired development repetitions allocate approximately
\(60\%/20\%/20\%\) of groups to fitting, stopping, and separate
configuration scoring. Each candidate's stopping-selected checkpoint
is scored in original response units on the configuration partition.
Each method independently minimizes the mean of these four scores
over rate and \(\beta\). A configuration is eligible only if all four development fits
succeed and give finite results. We break ties in favor of the smaller
rate exponent, then the smaller \(\beta\).

Initial controller scales are \(2^j\), \(j=-8,\ldots,-2\), for
LoopAda and the pooling control, with allowed bounds \(-12\leq j\leq2\).
Initial Adam endpoint rates are \(0.01\,2^j\), \(j=-4,\ldots,2\),
with bounds \(-10\leq j\leq6\).
Every rate is crossed with all three block exponents.
Within each dataset and outer fold, if a method attains its lowest
development MSE at a grid boundary, we extend every method's grid by
one exponent in that direction wherever its bounds permit. We repeat
this procedure until no further extension is indicated or possible.
All 60 rate selections were interior to the searched ranges.

With hyperparameters fixed, we evaluate each of the three methods in
four fresh paired repetitions per outer fold. Each repetition allocates
approximately \(80\%\) of the outer-training groups to fitting and
\(20\%\) to stopping. Standardization and checkpoint selection are
repeated within each fit. Each observation is held out once per
repetition. Its contribution to test MSE is the mean of its four
squared out-of-fold errors.

\subsubsection{Risk interpretation and intervals}

Under the regression model in
Section~\ref{sec:prediction-risk},
\[
 \E\{Y-f_w(X)\}^2
 =2R(w)+\E\{\operatorname{Var}(Y\mid X)\}.
\]
Differences preserve excess-risk ordering for a common distribution;
MSE ratios retain irreducible response variance.

We use 50,000 paired bootstrap resamples of groups with common group
multiplicities across methods. Sampled observation errors are summed
and divided by the sampled observation count, preserving equal
observation weights despite unequal group sizes.
For each comparison, the 2.5th and 97.5th percentiles of the
bootstrap MSE ratios give a pointwise 95\% interval.

The intervals are conditional on the observed groups and fitted
out-of-fold predictors. They omit variation from refitting, tuning,
and checkpoint selection, as well as dependence between groups or
between fits with overlapping training sets.

\subsubsection{Prediction results}

Table~\ref{tab:real-regression-detail} reports test MSE for all three
methods. For Portuguese grades, both pointwise intervals for the MSE
ratio lie below one. For CT slices, only the interval relative to the
pooling control lies below one. Both intervals include one for
Communities and Crime and for diabetes (Table~\ref{tab:real-regression-ratios}).

\begin{table}[htbp]
\centering\small
\caption{Prediction MSE in four cases}
\label{tab:real-regression-detail}
\setlength{\tabcolsep}{4pt}
\setbox0=\hbox{%
\begin{tabular}{@{}lrrr@{}}
\toprule
Case & LoopAda & \shortstack[r]{Pooling\\control} & Adam\\
\midrule
Student Portuguese & $2.7462678$ & $2.8808005$ & $2.9197682$\\
CT Slices & $46.444895$ & $49.731424$ & $50.492912$\\
Communities Crime & $0.024227075$ & $0.024341114$ & $0.024513654$\\
Diabetes & $3200.8141$ & $3242.1675$ & $3198.4647$\\
\bottomrule
\end{tabular}%
}
\noindent\makebox[\linewidth][c]{\ifdim\wd0>\linewidth\resizebox{\linewidth}{!}{\box0}\else\box0\fi}
\par\smallskip
\begin{minipage}{\linewidth}\footnotesize
All cases use \(T=8\), original response units, and equal observation
weights. Each method contributes 20 selected fits, five outer folds
times four evaluation repetitions. Squared errors, not predictions,
are averaged across repetitions. The pooling control is tuned separately.
\end{minipage}
\end{table}

\begin{table}[htbp]
\centering\small
\caption{Paired MSE ratios}
\label{tab:real-regression-ratios}
\setlength{\tabcolsep}{4pt}
\setbox0=\hbox{%
\begin{tabular}{@{}llrr@{}}
\toprule
Case & Comparator & Ratio & Pointwise 95\% interval\\
\midrule
Student Portuguese & Pooling control & $0.95330$ & $[0.92312,\,0.97823]$\\
 & Adam & $0.94058$ & $[0.88688,\,0.99045]$\\
\addlinespace[2pt]
CT Slices & Pooling control & $0.93391$ & $[0.88228,\,0.99852]$\\
 & Adam & $0.91983$ & $[0.83400,\,1.01891]$\\
\addlinespace[2pt]
Communities Crime & Pooling control & $0.99531$ & $[0.96413,\,1.03538]$\\
 & Adam & $0.98831$ & $[0.93126,\,1.05360]$\\
\addlinespace[2pt]
Diabetes & Pooling control & $0.98725$ & $[0.96539,\,1.00865]$\\
 & Adam & $1.00073$ & $[0.97774,\,1.02597]$\\
\bottomrule
\end{tabular}%
}
\noindent\makebox[\linewidth][c]{\ifdim\wd0>\linewidth\resizebox{\linewidth}{!}{\box0}\else\box0\fi}
\par\smallskip
\begin{minipage}{\linewidth}\footnotesize
Ratio \(=\mathrm{MSE}_{\rm LoopAda}/\mathrm{MSE}_{\rm comparator}\).
Each interval uses 50,000 paired group-bootstrap resamples and
conditions on the observed groups and fitted out-of-fold predictors.
\end{minipage}
\end{table}

\FloatBarrier